\documentclass[10pt,twocolumn,letterpaper]{article}

\usepackage[pagenumbers]{iccv} 
\usepackage{amssymb}
\usepackage{amsmath}
\usepackage{bbold}
\usepackage{bbm}
\usepackage{multirow}
\usepackage{pifont}
\usepackage{stfloats}
\usepackage{colortbl}
\usepackage{graphicx}
\usepackage{float}

\renewcommand{\paragraph}[1]{\vspace{1.25mm}\noindent\textbf{#1}}

\definecolor{mygray}{gray}{.92}
\definecolor{darkgreen}{rgb}{0.13, 0.55, 0.13}

\definecolor{my_blue}{HTML}{d3eaf2}
\definecolor{my_green}{HTML}{d7ecdc}
\definecolor{my_blue_green}{HTML}{D5EBE7}
\definecolor{c1}{HTML}{fb9f3a}
\definecolor{c2}{HTML}{bc3785}
\definecolor{c3}{HTML}{ed7953}
\definecolor{c4}{HTML}{46039f}
\def\OURS{PartGCD}
\usepackage[ruled,vlined]{algorithm2e}
\usepackage{algorithmic,algorithm2e,float}
\SetKwInput{KwInput}{Input}
\SetKwInput{KwOutput}{Output}

\def\ie{\emph{i.e.}}

\newcommand{\cmark}{\ding{51}}%
\newcommand{\xmark}{\ding{55}}%

\definecolor{iccvblue}{rgb}{0.21,0.49,0.74}
\usepackage[pagebackref,breaklinks,colorlinks,citecolor=iccvblue]{hyperref}

\title{Learning Part Knowledge to Facilitate Category Understanding for Fine-Grained Generalized Category Discovery}

\author{Enguang Wang$^1$, Zhimao Peng$^1$, Zhengyuan Xie$^1$, Haori Lu$^1$, Fei Yang$^{2,1}$, Xialei Liu$^{2,1}$\\
$^1$VCIP, CS, Nankai University \qquad $^2$NKIARI, Shenzhen Futian\\ 
{\tt\small \{enguangwang,zhimao796,xiezhengyuan,luhaori\}@mail.nankai.edu.cn} \\
{\tt\small \{feiyang,xialei\}@nankai.edu.cn}
}

\begin{document}
\maketitle
\begin{abstract}

Generalized Category Discovery (GCD) aims to classify unlabeled data containing both seen and novel categories.
Although existing methods perform well on generic datasets, they struggle in fine-grained scenarios.
We attribute this difficulty to their reliance on contrastive learning over global image features to automatically capture discriminative cues,  which fails to capture the subtle local differences essential for distinguishing fine-grained categories.
Therefore, in this paper, we propose incorporating part knowledge to address fine-grained GCD, which introduces two key challenges: the absence of annotations for novel classes complicates the extraction of the part features, and global contrastive learning prioritizes holistic feature invariance, inadvertently suppressing discriminative local part patterns.
To address these challenges, we propose \textbf{PartGCD}, including 1) Adaptive Part Decomposition, which automatically extracts class-specific semantic parts via Gaussian Mixture Models, and 2) Part Discrepancy Regularization, enforcing explicit separation between part features to amplify fine-grained local part distinctions.
 Experiments demonstrate state-of-the-art performance across multiple fine-grained benchmarks while maintaining competitiveness on generic datasets, validating the effectiveness and robustness of our approach.
\end{abstract}    
\section{Introduction}
\label{sec:intro}

With the increasing quantity of annotated data~\cite{imagnet}, image classification models~\cite{he2016deep,krizhevsky2017imagenet} have achieved remarkable advancements.
However, such supervised models are typically under a closed-set assumption, recognizing only the classes present in the training set, which forces any unseen classes to be misclassified as known categories, thus limiting their applicability in real-world settings. 
Generalized Category Discovery (GCD)~\cite{gcd} addresses this by jointly recognizing base and novel classes in unlabeled data.
While existing GCD methods show encouraging results, especially with stronger backbone architectures~\cite{zheng2024textual,wang2024get}, 
their performance gains primarily emerge on generic datasets like ImageNet-100, and remain less satisfactory on fine-grained datasets.
Meanwhile, studying category discovery on fine-grained datasets is quite essential, as it reduces the dependency on domain experts and enhances the scalability of category discovery systems. 
Therefore, in this paper, we focus on fine-grained GCD, a task that few works have attempted to address~\cite{xcon, infosieve, selex}.

\begin{figure}[t]
    \centering
    \includegraphics[width=\columnwidth]{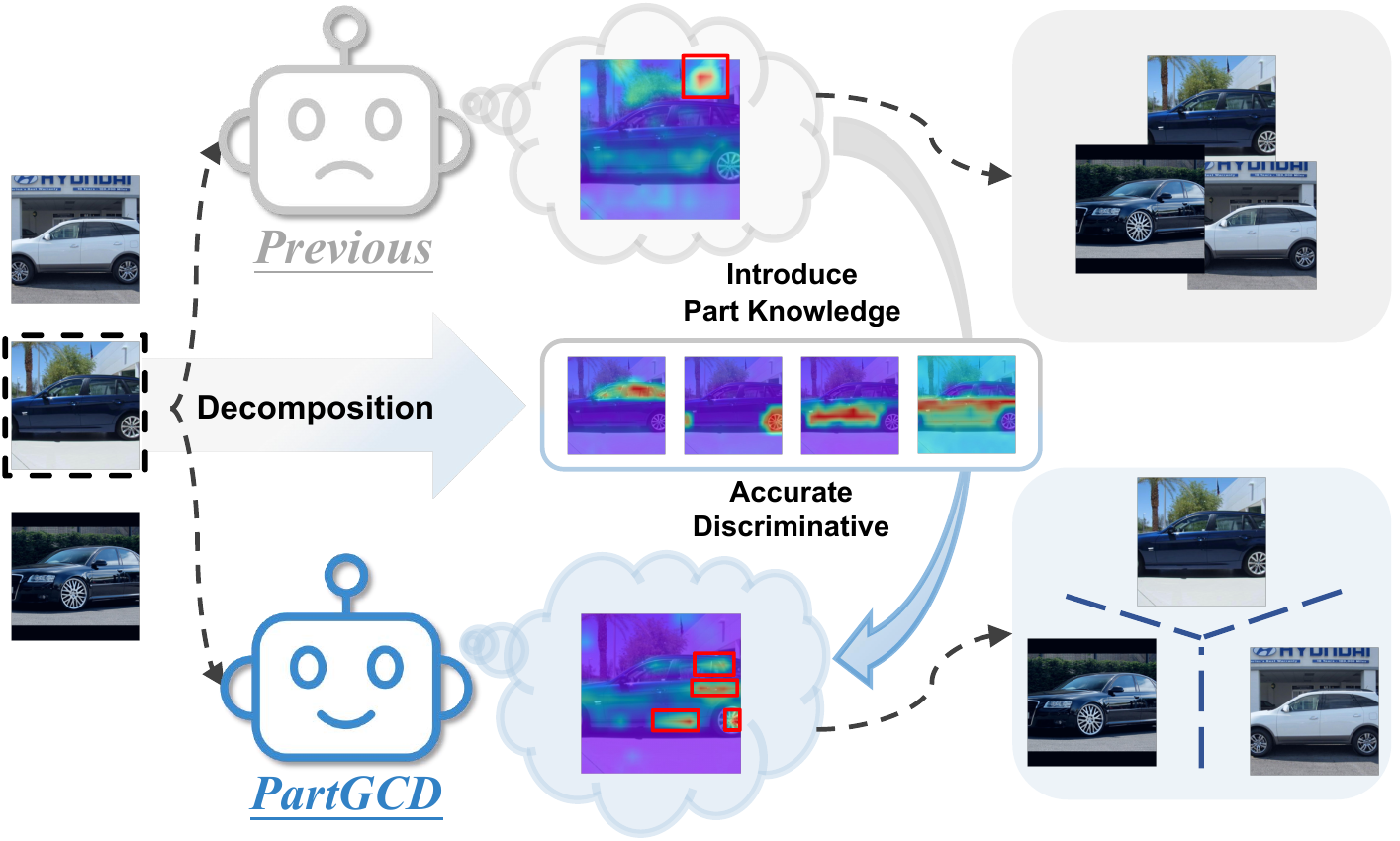 }
    \vspace{-6.5mm}
    \caption{
    Previous methods rely on global features, leading to inaccurate discrimination cues.
    Our PartGCD captures distinct part information, achieving accurate and discriminative perception.
    }
    \label{fig:teaser}
    \vspace{-7mm}
\end{figure}

The core of GCD lies in establishing visual-semantic relationships, which is challenging to capture due to the presence of unlabeled data.
Current GCD approaches~\cite{simgcd, mugcd, sptnet, legogcd, gcd, promptcal, dccl, gpc, xcon, infosieve, cms} commonly leverage instance-level contrastive learning on global image features to facilitate this relationship capture.
However, this learning strategy falls short on fine-grained datasets~\cite{cole2022does}, where subtle inter-class visual variations blur semantic boundaries, requiring the model to pay more attention to detailed differences within images to achieve effective differentiation. 
As illustrated in~\cref{fig:teaser}, this limitation results in discriminative regions, automatically captured from global representations, often mistakenly emphasizing background elements or inaccurate discrimination cues. 
This shortcoming highlights the limited capability of the model in category understanding, thereby hindering the establishment of robust visual-semantic relationships.
Moreover, we observe that while some categories (\eg, car models) are challenging to recognize from a global view, they exhibit distinguishing part features (\eg,  grille design, headlight shape).

Motivated by these analyses, we propose that incorporating part information into GCD learning can alleviate semantic ambiguities arising from subtle inter-class differences by facilitating the model’s category understanding. This idea is conceptually straightforward, but it addresses a critical gap in current approaches, which use contrastive learning on global features and, therefore, struggle with fine-grained distinctions. Meanwhile, it is also grounded in cognitive science findings~\cite{EDWARDS201921,10.5555/1854509.1854762} that indicate humans naturally decompose objects into parts for effective identification.

However, introducing part knowledge presents unique challenges in the GCD context: 
\textbf{the absence of labeled data for novel categories makes it challenging to extract part features, 
and global contrastive learning in GCD methods often reduce the model’s sensitivity to subtle inter-class differences essential for fine-grained perception}~\cite{cole2022does, park2023ssl}. 
To tackle these challenges, we propose \textbf{PartGCD}, including an Adaptive Part Decomposition (APD) strategy to guide the extraction of part features and a Part Discrepancy Regularization (PDR) to enhance the model’s local awareness.
Specifically, our APD begins by selecting pure samples, \ie, samples belonging to the same class, based on the distance between instance features and calibrated class prototypes, establishing the data foundation for extracting part features for novel classes. Using these selected samples, we then construct a part-based Gaussian Mixture Model (GMM) for each class,  where each Gaussian component represents the distribution of a specific part.
The part-based GMM then generates part attention maps, guiding the image patch features to form different semantic part features. 
Our PDR employs a contrastive objective on different part features within each image, encouraging clear semantic separation among part features and addressing the limitations of prior GCD methods that only apply objectives at the instance or cluster level, thus enhancing the model’s ability to capture fine-grained part-specific details.
Additionally, the model learns to adaptively aggregate parts, encouraging distinct response patterns across categories, thus improving its ability to handle fine-grained variations.

To summarize, our contributions are three-fold: 
\begin{itemize}
    \item  To address the challenge of extracting part features in the absence of annotations for novel classes, we propose an Adaptive Part Decomposition strategy to guide the part feature extraction process.
    \item 
    We propose a Part Discrepancy Regularization (PDR) to ensure clear semantic separation among part features, thus enhancing the model’s local awareness.
    \item Our PartGCD achieves significant performance improvement on multiple fine-grained datasets while maintaining competitiveness on generic datasets, emphasizing the importance of part knowledge in the GCD task. 
\end{itemize}

\section{Related Work}
\label{sec:rework}

\paragraph{Novel Class Discovery (NCD).}
The goal of NCD is to semantically cluster unlabeled novel data, given labeled data of known classes. This task was first proposed in~\cite{hsu2017learning, hsu2019multi} and formally defined in~\cite{han2019learning}. Earlier works mainly generated pseudo-labels for unlabeled samples by calculating the pairwise similarity between two samples~\cite{rs,zhong2021openmix,Zhong_2021_CVPR,zhao21novel}. Recently, the self-labeling~\cite{asano2020self} strategy has been introduced into the NCD task~\cite{uno, yang2022divide, li2023modeling, gu2023class} by assuming the novel data are equally partitioned into different classes, and generates pseudo-labels for unlabeled samples through solving the optimal transport problem.  

\paragraph{Generalized Category Discovery (GCD).}  
As an extension of NCD, GCD assumes that unlabeled data to be clustered contain both known and unknown classes~\cite{gcd}. 
Existing GCD methods can be classified into non-parametric~\cite{gcd, promptcal, dccl, gpc, xcon, infosieve, cms} and parametric approaches~\cite{simgcd, mugcd, sptnet, legogcd, peng2024let}. 
For the non-parametric GCD method, GCD~\cite{gcd} proposes fine-tuning the model with supervised contrastive learning on labeled data and self-supervised contrastive learning on the entire dataset. The discriminative representations of the training data are then clustered using semi-supervised k-means to obtain the final clustering results.  
For the parametric method, SimGCD~\cite{simgcd}  generates soft pseudo-labels using a self-distillation technique~\cite{dino} and employs a mean entropy maximization regularization term to prevent trivial solutions. 
Although previous methods have demonstrated promising results, most are more effective on generic datasets. Only a few works have focused on the fine-grained GCD.
XCon~\cite{xcon} first clusters the dataset into finer subclasses containing class-agnostic cues, and then applies contrastive learning on these subclasses, enabling the model to capture class-relevant fine-grained features. 
InfoSieve~\cite{infosieve} learns an implicit category code tree, framing the classification task as the search for a class code, resulting in significant improvements on fine-grained datasets.
SelEx~\cite{selex} generates pseudo-labels across different levels of class granularity, which dynamically guide the negative sample weights in unsupervised contrastive learning and provide supervision signals for supervised contrastive learning in subsequent epochs.
In contrast to these methods, we innovatively introduce part knowledge to address GCD.

\paragraph{Part-based Learning.} 
Part knowledge learning, by capturing part information from images as a complement to global features, effectively addresses the challenge of distinguishing similar categories. This approach has been extensively explored across various tasks, particularly in the fine-grained image recognition task~\cite{wei2021fine}. 
Early works~\cite{zhang2016spda,huang2016part,lin2015deep,zhang2014part,xiao2015application,wei2016mask} in fine-grained image recognition provide valuable local parts through approaches that leverage intrinsic model characteristics or utilize part annotations help models accurately discern fine-grained distinctions, thus improving classification accuracy.
Recent works~\cite{huang2020interpretable,chen2019looks,wang2021interpretable,donnelly2022deformable} capture category-specific part features through learnable prototypes for each category, achieving better results while providing interpretability. 
MGProto~\cite{wang2023mixture}
captures the class-conditional data density through GMMs to acquire more prototypical representation power,  DF-GMM~\cite{wang2020weakly} use GMMS to help extract discriminative regions, while they need image-level annotations.
To our best knowledge, we are the first to introduce part knowledge into GCD by explicit modeling of part features. Our work identifies the challenges of leveraging part knowledge in GCD and proposes effective solutions to address them.

\section{Method}

\begin{figure*}[t]
    \centering
    \includegraphics[width=1\textwidth]{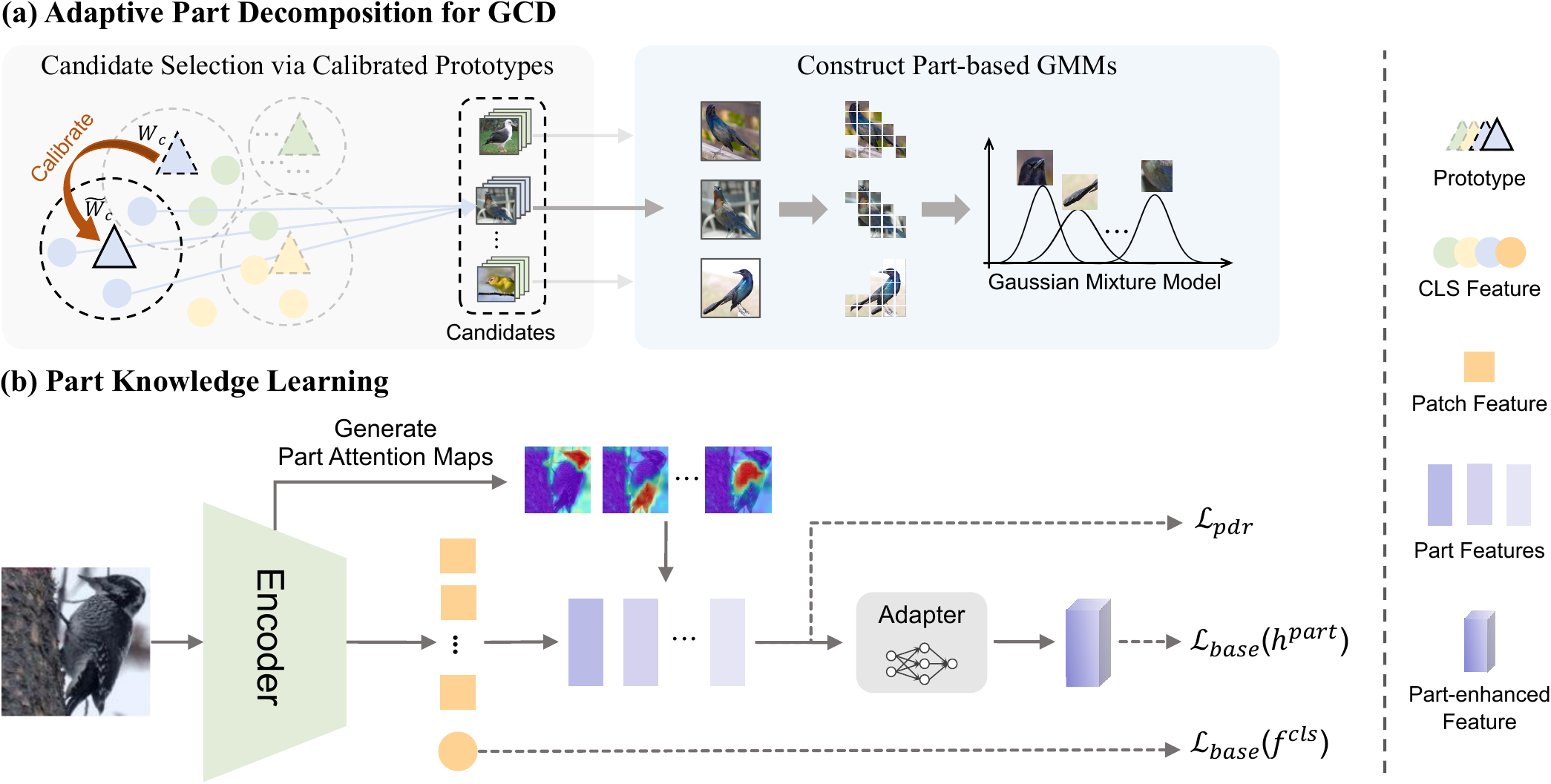}
    \caption{The framework of our PartGCD. At the beginning of each epoch, we select candidate samples for each class via calibrated prototypes and then use these samples to construct part-based GMMs, assigning each sample corresponding GMM parameters. 
    At each iteration,
    we generate part attention maps to guide part knowledge learning.   
    }
    \label{fig:pipline}
    \vspace{-5mm}
\end{figure*}

In this section, we propose  \textbf{PartGCD}, 
a novel approach designed to incorporate part knowledge into  GCD for robust category discovery. The overall framework of our method is depicted in \cref{fig:pipline}. Below, we first revisit the preliminaries before elaborating on our method.

\subsection{Preliminaries}
\paragraph{Problem Formulation.}
Let~$\mathcal{D}_l=\left\{(\boldsymbol{x}_i, \boldsymbol{y}_i)\right\}\subset \mathcal{X}\times \mathcal{Y}_l$ and $\mathcal{D}_u=\left\{(\boldsymbol{x}_i, \boldsymbol{y}_i)\right\} \subset \mathcal{X}\times \mathcal{Y}_u$
represent the labeled and unlabeled datasets, respectively. 
$\mathcal{Y}_l$ and $\mathcal{Y}_u$ denote the corresponding label spaces, with $\mathcal{Y}_l \subset \mathcal{Y}_u$, thus the label space of new classes in unlabelled data is $\mathcal{Y}_n = \mathcal{Y}_u\backslash\mathcal{Y}_l$. The complete training set is $\mathcal{D} = \mathcal{D}_l \cup \mathcal{D}_u $. 
Consistent with prior works~\cite{simgcd,promptcal,legogcd}, we assume the class number of new classes $|\mathcal{Y}_n|$, is known or it can be estimated through some off-the-shelf methods~\cite{gcd,han2019learning}.
The total number of classes is denoted by $C = |\mathcal{Y}_u|$. The target of the GCD task is to classify the samples in the unlabeled dataset $\mathcal{D}_u$.

\paragraph{Parametric Baseline.} 
Given an image $\boldsymbol{x}$,  a DINO~\cite{dino} pre-trained Vision Transformer (ViT)  is used to extract its feature representations [$\boldsymbol{f}^{cls}$, 
$\boldsymbol{f}^{1}$, ..., $\boldsymbol{f}^{N_p}$], where $N_p$ is the total number of patch tokens, $\boldsymbol{f}^{cls}$ and [$\boldsymbol{f}^{1}$, ..., $\boldsymbol{f}^{Np}$] correspond to the class token and the patch tokens, respectively. 
The parametric baseline~\cite{simgcd} performs both contrastive representation learning and classification learning on $\boldsymbol{f}^{cls}$. 
For classification learning,  a cross-entropy loss $\ell$ is applied to the labeled subset $B_l$ of the mini-batch $B$ :
\begin{equation} 
\mathcal{L}^{s}_\text{cls}(\boldsymbol{f}^{cls}) = \frac{1}{|B_l|}\sum_{\boldsymbol{x}_i \in B_l}{\ell\left(\boldsymbol{y}_i,\boldsymbol{p}_i\right)},\,
\end{equation}
where
$\boldsymbol{p}_i = \delta(
\frac{\boldsymbol{W}^\top\boldsymbol{f}^{cls}_i }{\tau_s}) \in \mathbb{R}^C$
represents the
prediction with a temperature parameter  $\tau_s$, $\delta$ is the softmax function. $\boldsymbol{W} \in \mathbb{R}^{dim \times C}$ denotes the class prototypes , where $dim$ represent feature dimension,  both $\boldsymbol{f}^{cls}_i$  and $\boldsymbol{W}$ are  $l_2$-normalized.
Moreover, a soft pseudo-label generated from another view $\boldsymbol{x}'$ supervises the current view using a sharper temperature $\tau_t$:
\begin{equation} 
\mathcal{L}^{u}_\text{cls}(\boldsymbol{f}^{cls}) = \frac{1}{|B|}\sum_{\boldsymbol{x}_i \in B}{\ell\left(\boldsymbol{p}'_i,\boldsymbol{p}_i \right)} \,.
\end{equation}

Thus, the overall classification loss can be expressed as $\mathcal{L}_\text{cls}(\boldsymbol{f}^{cls}) =  \lambda \mathcal{L}^{s}_\text{cls}(\boldsymbol{f}^{cls}) + (1-\lambda)\mathcal{L}^{u}_\text{cls}(\boldsymbol{f}^{cls}) + \zeta $,
where $\zeta$ is a mean-entropy maximization regularization term~\cite{assran2022masked} to prevent trivial solutions, and $\lambda$ is a balance hyperparameter. Given that contrastive representation learning loss $\mathcal{L}_\text{rep}(\boldsymbol{f}^{cls})$  is commonly employed in GCD methods, we provide its expression in the \textit{supplementary materials (supp)}.  
The overall objective for the parametric baseline is:
\begin{equation} 
\label{simgcd_loss}
\mathcal{L}_\text{base}(\boldsymbol{f}^{cls}) =  \mathcal{L}_\text{cls}(\boldsymbol{f}^{cls}) + \mathcal{L}_\text{rep}(\boldsymbol{f}^{cls}) \,.
\end{equation}

While previous methods show promising performance, they fall short in establishing robust visual-semantic relationships for fine-grained images arising from their reliance on global features alone, which omits the distinct semantics in local features.  
Additionally, global contrastive learning, though effective for instance-level discrimination, often overlooks fine-grained details, further leading to suboptimal results on fine-grained data. 
To address these limitations, we introduce part knowledge into the learning process, enhancing the model's fine-grained category discovery ability.

\subsection{Adaptive Part Decomposition for GCD}
\label{ssec:apd}

The lack of annotations for new classes poses a unique challenge in extracting part features in the GCD task. 
To tackle this, in our APD,  we first select high-purity candidate samples for novel classes based on calibrated class prototypes. These candidates enable the construction of part-based Gaussian Mixture Models, which generate part attention maps to guide the part extraction.

\paragraph{Candidate Selection via Calibrated Prototypes.}
Motivated by the observation that clustering accuracy for new classes improves as training progresses~\cite{simgcd}, we incorporate model predictions into the sample selection process.
A straightforward method is to select a fixed number of samples per class based on model predictions.
Given model predictions  $\boldsymbol{P} = [\boldsymbol{p}_1;...;\boldsymbol{p}_{|\mathcal{D}|}] \in \mathbb{R}^{|\mathcal{D}|\times C} $, the indices of the selected candidate samples for the $c$-th class are:
\begin{equation}
\label{eq4}
\mathcal{A}_c = \mathrm{arg\,top}^{N_s}_i\,\boldsymbol{P}_{i,c} \,,
\end{equation}
where $\mathrm{arg\,top}^{N_s}$ returns the indices of the top-${N_s}$ items. 

However, due to the strong supervision of old classes, the prototypes of new classes may be biased toward old classes during early training stages, reflecting the classifier's insufficient learning and 
leading to noisy selection (\ie, candidate samples with low purity).
To mitigate this issue, inspired by optimal transport-based clustering~\cite{asano2020self}, we impose a uniform constraint on the predictions by setting $\sum_{i}^{|\mathcal{D}|}{p^{(c)}_i} = \frac{|\mathcal{D}|}{C}$ for $c = 1, ..., C$, yielding the adjusted predictions $[\boldsymbol{q}_1;...;\boldsymbol{q}_{|\mathcal{D}|}] = \text{SK}([\boldsymbol{p}_1;...;\boldsymbol{p}_{|\mathcal{D}|}])$, where SK($\cdot$) represents the Sinkhorn-Knopp algorithm~\cite{sinkhorn}, an efficient method for solving optimal transport problems. 
We provide its theoretical justification in \textit{supp}.

The adjusted predictions are then used to calibrate the class prototype:
\begin{equation}
\label{eq5}
    \Tilde{\boldsymbol{W}}_c =  \frac{\sum_{i\in \mathcal{Q}_c } \boldsymbol{q}^{(c)}_i \cdot \boldsymbol{f}^{cls}_i}{\sum_{j\in \mathcal{Q}_c}\boldsymbol{q}^{(c)}_j}  ,
\end{equation}
where $\mathcal{Q}_c$ is the set of sample indices predicted to class $c$ based on the adjusted predictions. 
We then refine the selection of candidate samples for each class as:
\begin{equation}
\label{eq:calibrated_candidates}
\Tilde{\mathcal{A}}_c = \mathrm{arg\,top}^{N_s}_i\,\Tilde{\boldsymbol{P}}_{i,c} \,,
\end{equation}
where $\Tilde{\boldsymbol{P}} = [\Tilde{\boldsymbol{p}}_1;...;\Tilde{\boldsymbol{p}}_{|\mathcal{D}|}] $, and $\Tilde{\boldsymbol{p}}_i = {\delta}(
\Tilde{\boldsymbol{W}}^\top\boldsymbol{f}^{cls}_i)$ is the prediction using calibrated class prototypes.
Since the data distribution of new classes is unknown, we set ${N_s}$  based on the average samples across the old classes, \ie, 
${N_s} = \gamma *\frac{|\mathcal{D}_l|}{|\mathcal{Y}_l|}$
where $\gamma$ is a hyperparameter. 
We select candidates for new classes and use labeled data as old classes's candidates.

\paragraph{Construct Part-based GMMs.}
The Gaussian Mixture Model (GMM) effectively captures data distribution through soft feature assignment, which makes it robust to noise and enhances part modeling accuracy—essential for our part-based approach. 
Therefore, leveraging the GMM's ability to model underlying distributions and reject noise, we fit a part-based GMM for each class to generate part attention maps.
To start with, we filter patches using attention values to  prevent background overfitting, thus the patch features to fit the GMMs for class $c$  are defined as:
\begin{equation}
\mathcal{P}_c = \bigcup_{i \in \Tilde{\mathcal{A}_c}} \{\boldsymbol{f}_i^j \mid {attn}_i^j \geq \text{Average}({attn}_i)\},
\end{equation}
where $\boldsymbol{f}^j$ represents the $j$-th patch feature and $attn_i \in \mathbb{R}^{N_p}$ is the class-to-patch token attention average across all heads in the last block which provides a strong observation of the foreground~\cite{dino}.
Since patch features from the DINO pre-trained ViT backbone contain local semantic information and align with object parts~\cite{amir2021deep},  multivariate Gaussians can represent different object parts effectively.
Following~\cite{liang2022gmmseg}, we formulate the GMM  as:
\begin{align}
\begin{split}
\label{eq:gmm}
p(\boldsymbol{f}^j|\boldsymbol{\xi}) = & \sum_{k=1}^{K}{{p({part}_{k}|\pi_k)} p(\boldsymbol{f}^j|{part}_{k};\boldsymbol{\mu}_k,\boldsymbol{\Sigma}_k)} \\
    = & \sum_{k=1}^{K}\pi_k{\mathcal{N}(\boldsymbol{f}^j;\boldsymbol{\mu}_k,\boldsymbol{\Sigma}_k})\,,
\end{split}
\end{align}
where  $\boldsymbol{\xi} = \{\pi_k, \boldsymbol{\mu}_k,\boldsymbol{\Sigma}_k\}_{k=1}^K$ are the parameters for the multivariate Gaussian~$\mathcal{N}$, 
${part}_{k}$ is the $k$-th part of the class, and $K$ is the number of parts that are consistent across all classes.
Conduct the standard EM algorithm on $\mathcal{P}_c$ can find the optimal parameters $\boldsymbol{\xi}^c$ for class $c$.
In our approach, $K$ is automatically determined using the Silhouette Coefficient~\cite{ROUSSEEUW198753},
a standard clustering metric that assesses cluster quality, a higher value indicates that a point is well matched to its cluster. 
Specifically, $K$ is chosen based on the average silhouette coefficient across the test data of old classes. We provide detailed calculations in \textit{supp}.

Given  GMM parameters $\{\boldsymbol{\xi}^c\}_{c=1}^C$ for all classes, we first assign the corresponding GMM parameters for a sample $\boldsymbol{x}_i$ as $\boldsymbol{\xi}^{a_i} = \{\pi_k^{a_i}, \boldsymbol{\mu}_k^{a_i},\boldsymbol{\Sigma}_k^{a_i}\}_{k=1}^K$, where $a_i ={\arg\max}_c \,  q^{(c)}_i$. According to the Bayes’ theorem and~\cref{eq:gmm}, the posterior probability for a patch feature $\boldsymbol{f}^{j}_i$ is calculated by:
\begin{align}
\label{eq9}
\begin{split}
    p(part_{k}|\boldsymbol{f}^{j}_i) = & \frac{p(\boldsymbol{f}^{j}_i|{part}_{k})p({part}_{k})}{\sum_{k'}p(\boldsymbol{f}^{j}_i|{part}_{k'})p({part}_{k'})} \\
    = &
 \frac{\pi_k^{a_i}{\mathcal{N}(\boldsymbol{f}^{j}_i;\boldsymbol{\mu}_k^{a_i},\boldsymbol{\Sigma}_k^{a_i}})}{\sum_{k'=1}^{K}\pi_{k'}^{a_i}{\mathcal{N}(\boldsymbol{f}^{j}_i;\boldsymbol{\mu}_{k'}^{a_i},\boldsymbol{\Sigma}_{k'}^{a_i}})} \,,
\end{split}
\end{align}
 indicating the probability that $\boldsymbol{f}^{j}_i$ belongs to $part_k$.
Based on this, we define the part attention map for image $\boldsymbol{x}_i$
on ${part}_{k}$ as $\boldsymbol{M}_{i}^{k} = [p({part}_{k}|\boldsymbol{f}^{1}_i);...;p({part}_{k}|\boldsymbol{f}^{N_p}_i) ] \in \mathbb{R}^{N_p}$.

\subsection{Part Knowledge Learning}

Another crucial challenge in introducing part knowledge into GCD is that commonly used global contrastive learning in previous methods prioritizes holistic feature invariance, inadvertently suppressing discriminative local part patterns.

In order to enhance the model’s ability to capture fine-grained part-specific details and improve its discrimination across parts, which is crucial for distinguishing fine-grained categories, we propose a  Part Discrepancy Regularization to ensure a clear semantic distance among part features. 
To be specific, given part attention map $\boldsymbol{M}_{i}^{k}$ for image $\boldsymbol{x}_i$, the $k$-th part feature $\boldsymbol{v}^{k}_i$ 
is calculated as:
\begin{equation}
\label{eq:part_f}
 \boldsymbol{v}^{k}_i = \sum_{j}\,\boldsymbol{M}_{i,j}^{k}\,\boldsymbol{f}^{j}_i\,.
\end{equation}

We then employ a contrastive loss to form our Part Discrepancy Regularization:
\begin{equation}
\label{eq:loss_pdr}
\mathcal{L}_{pdr}  = -\frac{1}{|B|K}\sum_{i \in B}\sum_{k \in {K}}{\text{log}{\frac{\text{exp}({\boldsymbol{v}^{k}_{i}}^\top \boldsymbol{v}^{k}_{i'})}{\sum_{k'\neq k}\text{exp}({\boldsymbol{v}^{k}_{i}}^\top \boldsymbol{v}^{k'}_{i'})}}},
\end{equation}
where $\boldsymbol{v}^{k}_{i'}$ is the part feature of another view $\boldsymbol{x'}_{i}$.
This regularization employs a contrastive objective to encourage distinct semantic representations for each part feature, effectively maximizing semantic separation among parts within the same object. This separation is essential for fine-grained category discovery as it prevents feature overlap between visually similar categories in a fine-grained perspective.

 Meanwhile, we employ an MLP  as an adapter to aggregate these part features, obtaining the part-enhanced feature $\boldsymbol{h}_i^{part}$ for $\boldsymbol{x}_i$  by $\boldsymbol{h}_i^{part} = \text{MLP}([\boldsymbol{v}^{1}_i,...,\boldsymbol{v}^{K}_i])$. Then,
conduct  the parametric training~\cref{simgcd_loss} on the part-enhanced features:
\begin{equation} 
\label{eq12}
\mathcal{L}_\text{base}(\boldsymbol{h}^{part}) =  \mathcal{L}_\text{cls}(\boldsymbol{h}^{part}) + \mathcal{L}_\text{rep}(\boldsymbol{h}^{part}) \,.
\end{equation} 
 Through contrastive representation learning and classification learning on the part-enhanced feature, 
 the adapter learns to selectively emphasize parts relevant to each class, tailoring the part-based representations to category-specific patterns,
 thus improving its ability to handle fine-grained variations that are essential for effective category discovery.
 
Together with the parametric training on the class token, the final objective of our PartGCD is formulated as follows:
\begin{equation} 
\label{eq13}
\mathcal{L}= \mathcal{L}_\text{base}(\boldsymbol{f}^{cls})  + \alpha \,(\mathcal{L}_\text{base}(\boldsymbol{h}^{part}) + \mathcal{L}_{pdr})\,,
\end{equation} 
where $\alpha$ balances the model's focus on local and global knowledge.
Through joint training, our approach enhances the model’s visual understanding of different categories, enabling a more refined and structured visual-semantic relationship, ultimately enabling a richer and more discriminative category differentiation in the GCD task, especially in challenging fine-grained datasets.

In our method, we perform sample selection and GMM construction only at the beginning of each epoch, assigning each sample corresponding GMM parameters. At each iteration within the epoch, we generate part attention maps for each mini-batch sample to guide part knowledge learning. We provide the pseudo-code of our method in the \textit{supp}.
The final class assignment of $\boldsymbol{x}_i$ integrates predictions derived from both its global and part information:
\begin{equation} 
\label{eq14}
\hat{y}_i = \mathrm{arg\,max}_c\,
{\delta}^{(c)}(
{\boldsymbol{W}^\top\boldsymbol{f}^{cls}_i }) + {\delta}^{(c)}(
{\boldsymbol{W}^\top\boldsymbol{h}^{part}_i }) \,. 
\end{equation}

\section{Experiments}

\subsection{Experimental setup
}
\paragraph{Datasets and evaluation.}
We evaluate our method on three generic datasets, four fine-grained datasets, and three ultra-fine-grained datasets.
 The statistics of the dataset splits and evaluation protocol can be found in \textit{supp}.

\paragraph{Implementation details.}
We use a DINO~\cite{dino} pretrained ViT-B/16 as the backbone, leading to $N_p$ = 196. 
$\alpha$ is set to 2 for all fine-grained datasets and 1 for all generic datasets.
$\gamma$ is set to 1. 
Based on the average silhouette coefficient across the test data of old classes, $K$ is set to 5 for CUB and Aircraft datasets, 6 for the SCars (short for SanfordCars) dataset, 3 for Cifar100 and Herb (short for Herbarium19) datasets, and 4 for Cifar10 and ImageNet100 datasets. We provide silhouette coefficient statistics and more implementation detail in \textit{supp}.

\begin{table*}[tp]
\small
\centering

\resizebox{\textwidth}{!}{%
\begin{tabular}{lccc|ccc|ccc|ccc|ccc}
\toprule
   & \multicolumn{3}{c}{Stanford Cars} & \multicolumn{3}{c}{FGVC-Aircraft} & \multicolumn{3}{c}{Herbarium19} &   \multicolumn{3}{c}{CUB} & \multicolumn{3}{c}{Average}
\\
\cmidrule(rl){2-4}\cmidrule(rl){5-7}\cmidrule(rl){8-10} \cmidrule(rl){11-13} \cmidrule(rl){14-16}
  Method    & All  & Old  & New  & All  & Old  & New  &  All & Old  & New &   All & Old  & New & All & Old  & New    \\
\cmidrule[0.5pt]{1-16} 
RS+~\cite{rs} & 28.3 & 61.8 & 12.1 & 26.9 & 36.4 & 22.2& 27.9 &55.8 &12.8  & 33.3 & 51.6 & 24.2& 29.1 & 51.4 & 17.8 \\ 
UNO+~\cite{uno}  & 35.5 & 70.5 & 18.6 & 40.3 & 56.4 & 32.2 &28.3& 53.7& 14.7& 35.1 & 49.0 & 28.1& 34.8 & 57.4 & 23.4 \\ 
ORCA~\cite{orca} & 31.9&42.2&26.9 &31.6&	32.0&	31.4 &  20.9 &30.9& 15.5& 36.3&	43.8&	32.6& 30.2 & 37.2 & 26.6 \\ 
GCD~\cite{gcd}       & {39.0} & 57.6 & {29.9} & {45.0} & 41.1 & {46.9}  &35.4 &51.0 &27.0 & {51.3} & {56.6} & {48.7}  & 50.2 & 58.3 & 46.0 \\ 
PromptCAL~\cite{promptcal}& {50.2} & 70.1 & {40.6} & {52.2} &52.2 & {52.3} &37.0& 52.0& 28.9 & {62.9} & {64.4} & {62.1} & 50.6 & 59.7 & 46.0 \\ 
DCCL~\cite{dccl} & {43.1} & 55.7 & {36.2} & {-} &- & {-} & -&-&-& {63.5} & {60.8} & {64.9}  & -&-&-\\ 
GPC~\cite{gpc}       & {42.8} & 59.2 & {32.8} & {46.3} & 42.5 & {47.9}&-& -&- & {55.4} & {58.2} & {53.1}&-& -&- \\ 
XCon~\cite{xcon}  & 40.5 & 58.8& 31.7&47.7 &44.4 &49.4& - & - & -& 52.1& 54.3 &51.0 & - & - & -\\ 
TIDA~\cite{tida} & 54.7 &72.3 &46.2 & 54.6 & 61.3 & 52.1& - & - & - & - & - & -&- & - & -\\ 
SimGCD~\cite{simgcd}     & 53.8 & 71.9 & 45.0 & 54.2 & 59.1 & 51.8 &44.0 & 58.0  &36.4 & 60.3 & 65.6 & 57.7& 53.1 & 63.6 & 47.7\\
CMS~\cite{cms} & 56.9 &76.1& 47.6 &56.0& 63.4& 52.3&36.4 &54.9 &26.4& 68.2 &\underline{76.5 }&64.0& 54.4 & 67.7 & 47.6\\ 
LegoGCD~\cite{legogcd} &57.3& 75.7 &48.4 &55.0& 61.5 &51.7&45.1& 57.4& \underline{38.4}& 63.8 &71.9 &59.8 & 55.3 & 66.6 & 49.6\\ 
SPTNet~\cite{sptnet} & \underline{59.0}& \underline{79.2}& 49.3 &\underline{59.3} &61.8 &\textbf{58.1} &43.4& 58.7 &35.2& 65.8& 68.8 &65.1& 56.9 & 67.1 & \underline{51.9}\\
$\mu$GCD~\cite{mugcd} &  56.5&  68.1 & \underline{50.9}&  53.8 & 55.4 & 53.0& \underline{45.8} & \textbf{61.9} & 37.2&  65.7 & 68.0&  64.6& 55.5 & 63.4 & 51.4 \\ 
InfoSieve~\cite{infosieve} &55.7 &74.8 &46.4 &56.3 &\underline{63.7} &52.5 &41.0 &55.4 &33.2 &\underline{69.4}& \textbf{77.9 }&65.2& 55.6 & \textbf{67.9} & 49.3\\ 
SelEx~\cite{selex}  &  58.5 & 75.6 & 50.3
&57.1&\textbf{ 64.7}& 53.3& 39.6 &54.9 &31.3& \textbf{73.6}& {75.3}& \textbf{72.8}&\underline{57.2} & 67.6 & \underline{51.9}\\ 
\midrule
\rowcolor{mygray}
\textbf{PartGCD (Ours)}  & \textbf{65.6} & \textbf{79.5} & \textbf{58.9} & \textbf{59.4} & {63.5} & \underline{57.4} &\textbf{46.1} & \underline{59.1} & \textbf{39.0}  & {68.6} & {68.9} & \underline{68.4}& \textbf{59.9} & \underline{67.8} & \textbf{55.9}\\
\bottomrule
\end{tabular}}
\vspace{-2mm}
\caption{{Results (\%) on fine-grained datasets.} Bold and underlined values represent the best and the second-best results, respectively.} 
\label{subtab:ssb}
\vspace{-5mm}
\end{table*}

\begin{table}[tp]
\small
\centering
\setlength{\tabcolsep}{2pt}
\resizebox{\columnwidth}{!}{%
    \begin{tabular}{lccc|ccc|ccc}
      \toprule
      \multirow{2}{*}{} & \multicolumn{3}{c}{SoyAgeing-R1} & \multicolumn{3}{c}{SoyAgeing-R3} & \multicolumn{3}{c}{SoyAgeing-R4}  \\
\cmidrule(rl){2-4}\cmidrule(rl){5-7}\cmidrule(rl){8-10} 
  Method    & All  & Old  & New  & All  & Old  & New  &  All & Old  & New    \\
      \midrule
      SimGCD$^\dag$~\cite{simgcd} & 44.0& 52.2& 39.9& 42.5& 52.0& 37.8& 43.8& 50.8& 40.3 \\
      InfoSieve$^\dag$~\cite{infosieve}  & 43.5& 49.9& 40.3& 41.0& 46.6& 38.1& 41.5& 44.8& 39.9 \\
      SelEX$^\dag$~\cite{infosieve} & 47.6& 53.6& \textbf{44.6} & 43.9& 49.3& 41.1& 43.8& 46.9& 42.2\\
      
      \midrule
      \rowcolor{mygray}
      \textbf{PartGCD} & \textbf{48.7} & \textbf{57.0} & 44.5 & \textbf{46.6} & \textbf{55.2} & \textbf{42.3} & \textbf{45.4} & \textbf{50.9} & \textbf{42.6}\\
        
      \bottomrule
    \end{tabular}
}
\vspace{-2mm}
\caption{Results (\%) on ultra-fine-grained datasets. $\dag$ denotes adapting the methods using the official code (DINO backbone).}
\label{tab:gcd_uf}
\vspace{-5mm}
\end{table}

\subsection{Comparison with the State-of-the-Art}

\paragraph{Comparison on fine-grained datasets.}
As shown in~\cref{subtab:ssb}, our method achieves leading performance on three of four fine-grained datasets, with the highest average accuracy—surpassing the previous best by 2.7\% on ``All'' and 4.0\% on ``New''. 
Besides, our method excels on the challenging, large-scale, and imbalanced Herb dataset, exceeding InfoSieve by 5.1\% and SelEX by 6.5\%, underscoring its effectiveness in leveraging fine-grained part features, even in complex, imbalanced scenarios.
We also observe a common performance bias among many methods. For example, SelEx and InfoSieve, which incorporate hierarchical structures into their design, achieve superior performance on CUB, a bird dataset with inherent hierarchical information. 
 As a result, our method falls behind them on this dataset, however, we outperform them on all other datasets.

Moreover, \cref{tab:gcd_uf} presents the comparisons on ultra-fine-grained datasets~\cite{Liu_2024_CVPR,yu2021benchmark}.
 These datasets' extremely minimal inter-class differences demand strong local perception from GCD models to capture subtle image variations. By incorporating local features into GCD learning, our method achieves significant improvements over other methods, demonstrating its robustness and effectiveness.

\begin{table}[tp]
\small
\centering

\setlength{\tabcolsep}{2pt}
\resizebox{\columnwidth}{!}{%
\begin{tabular}{lccc|ccc|ccc|c}
\toprule
   &    \multicolumn{3}{c}{CIFAR10} & \multicolumn{3}{c}{CIFAR100} & \multicolumn{3}{c}{ImageNet-100}& {Avg.}
\\
\cmidrule(rl){2-4}\cmidrule(rl){5-7}\cmidrule(rl){8-10}\cmidrule(rl){11-11}
   Method   & All  & Old  & New  & All  & Old  & New  & All  & Old  & New & All   \\
\cmidrule[0.5pt]{1-11} 
GCD      & {91.5} & {97.9} & {88.2} & {73.0} & 76.2 & {66.5} & {74.1} & 89.8 & 66.3 & 79.5\\ 
GPC   & {92.2} & \textbf{98.2} & { 89.1} & {77.9} & 85.0 & {63.0} & {76.9} & 94.3 & { 71.0} & 82.3 \\ 
XCon& 96.0& 97.3 &95.4 &74.2& 81.2& 60.3 &77.6& 93.5 &69.7&     82.6  \\ 
TIDA &\textbf{ 98.2} & 97.9 & \underline{98.5} &\underline{82.3} & 83.8  & \underline{80.7 } & - & - & - &  -  \\ 
SimGCD                  & 97.1 & 95.1 & 98.1 & 80.1 & 81.2 & {77.8} & 83.0 & 93.1 & 77.9 & 86.7  \\ 
CMS & - & - & - & \underline{82.3} &\textbf{ 85.7} &75.5 &84.7 & \textbf{95.6} &79.2 & - \\ 
LegoGCD&97.1& 94.3& \underline{98.5}& 81.8 &81.4 &\textbf{82.5}& \textbf{86.3} & \underline{94.5} &\textbf{ 82.1} &\underline{88.4} \\ 
SPTNet & \underline{97.3} &95.0& \textbf{98.6}& 81.3& 84.3 &75.6 & {85.4} &93.2& 81.4 & 88.0 \\ 
InfoSieve& 94.8& 97.7 &93.4& 78.3 &82.2 &70.5 &80.5& 93.8 & 73.8 &84.5  \\ 
SelEx & 95.9& \underline{98.1}& 94.8& \underline{82.3}& \underline{ 85.3} &76.3 &83.1 &93.6 &77.8&87.1  \\ 

\midrule
\rowcolor{mygray}
\textbf{PartGCD}  &  \underline{97.3} & {95.7} & {98.1} & \textbf{82.4} &{84.1} & {79.0} & \underline{85.8} & {94.4} &  \underline{81.5} &  \textbf{88.5} \\
\bottomrule
\end{tabular}}
\vspace{-2mm}
\caption{{Results (\%) on generic datasets.} }
\label{subtab:gen}
\vspace{-7mm}
\end{table}

\paragraph{Comparison on generic datasets.}
\cref{subtab:gen} shows the performance on generic datasets. While approaches like LegoGCD and SPTNet are effective in generic datasets, our method outperforms LegoGCD on CIFAR-10 and CIFAR-100, and  outperforms SPTNet on CIFAR-100 and ImageNet-100. 
Besides, our method surpasses fine-grained methods, such as SelEx, across all three generic datasets, underscoring its versatility. This adaptability stems from the ability to distinguish visually similar classes, which is beneficial in generic data (refer to~\cref{qr}). Moreover, our method shows a better efficiency than previous (in \textit{supp}).

Overall, our method, designed for fine-grained GCD tasks, demonstrates strong performance on both balanced/imbalanced/ultra fine-grained datasets, and also achieves competitiveness on generic datasets. This highlights the importance of incorporating part-based knowledge into GCD learning.

\begin{table}[!t]
\small
\centering
\setlength{\tabcolsep}{2.8pt}
  \resizebox{\columnwidth}{!}{
  \begin{tabular}{lcccccccccccc} \toprule
    \multirow{2}{*}{} & \multirow{2}{*}{$\mathcal{L}_{pdr}$} & \multicolumn{3}{c}{$\mathcal{L}_\text{base}^{part}$}&  \multicolumn{3}{c}{SCars} & \multicolumn{3}{c}{CUB} \\ 
    \cmidrule(rl){3-5} \cmidrule(rl){6-8} \cmidrule(rl){9-11}
    & & \multicolumn{1}{c}{$\mathcal{L}_\text{cls}^{part}$}  & \multicolumn{1}{c}{$\mathcal{L}_\text{rep}^{part}$}  & $\boldsymbol{h}^{part}$ & \multicolumn{1}{c}{All} & \multicolumn{1}{c}{Old}  & New & \multicolumn{1}{c}{All} & \multicolumn{1}{c}{Old}  & New  \\ \midrule
    \multicolumn{1}{c}{(1)} & \xmark & \xmark & \xmark  & \xmark & 53.8 & 71.9 & 45.0 &60.3& 65.6 &57.7  \\ 
    \multicolumn{1}{c}{(2)} & \cmark & \xmark & \xmark &  \xmark &55.1 & 72.5 & 46.6 &63.4&67.1&61.6\\
    \multicolumn{1}{c}{(3)} & \cmark & \cmark & \xmark &  \cmark & 61.8& 75.5 & 55.1& 66.9&68.1&66.4 \\  
    \multicolumn{1}{c}{(4)} & \cmark & \cmark & \cmark &  \xmark & 57.7 &	73.8 &	50.0 & 65.4&\textbf{69.8}&63.2\\  
     \multicolumn{1}{c}{(5)} & \xmark & \cmark & \cmark &  \cmark  & 62.0&76.3&55.1 & 63.8& 66.6& 62.4\\
     \rowcolor{mygray}
    \multicolumn{1}{c}{(6)} & \cmark & \cmark & \cmark &  \cmark& \textbf{65.6} & \textbf{79.5} & \textbf{58.9}& \textbf{68.6} & {68.9}& \textbf{68.4}\\ 
    \bottomrule
\end{tabular}}
\vspace{-3mm}
\caption{{Ablation study of main components.} Without $\boldsymbol{h}^{part}$ 
indicates using the average of part features without using the adapter.
}
\label{tab:diff_com}
\vspace{-5mm}
\end{table}

\subsection{Ablation and Analysis}
\label{ssec:ablations}

\begin{figure}[!h]
    \centering
    \vspace{-4mm}
\includegraphics[width=\columnwidth]{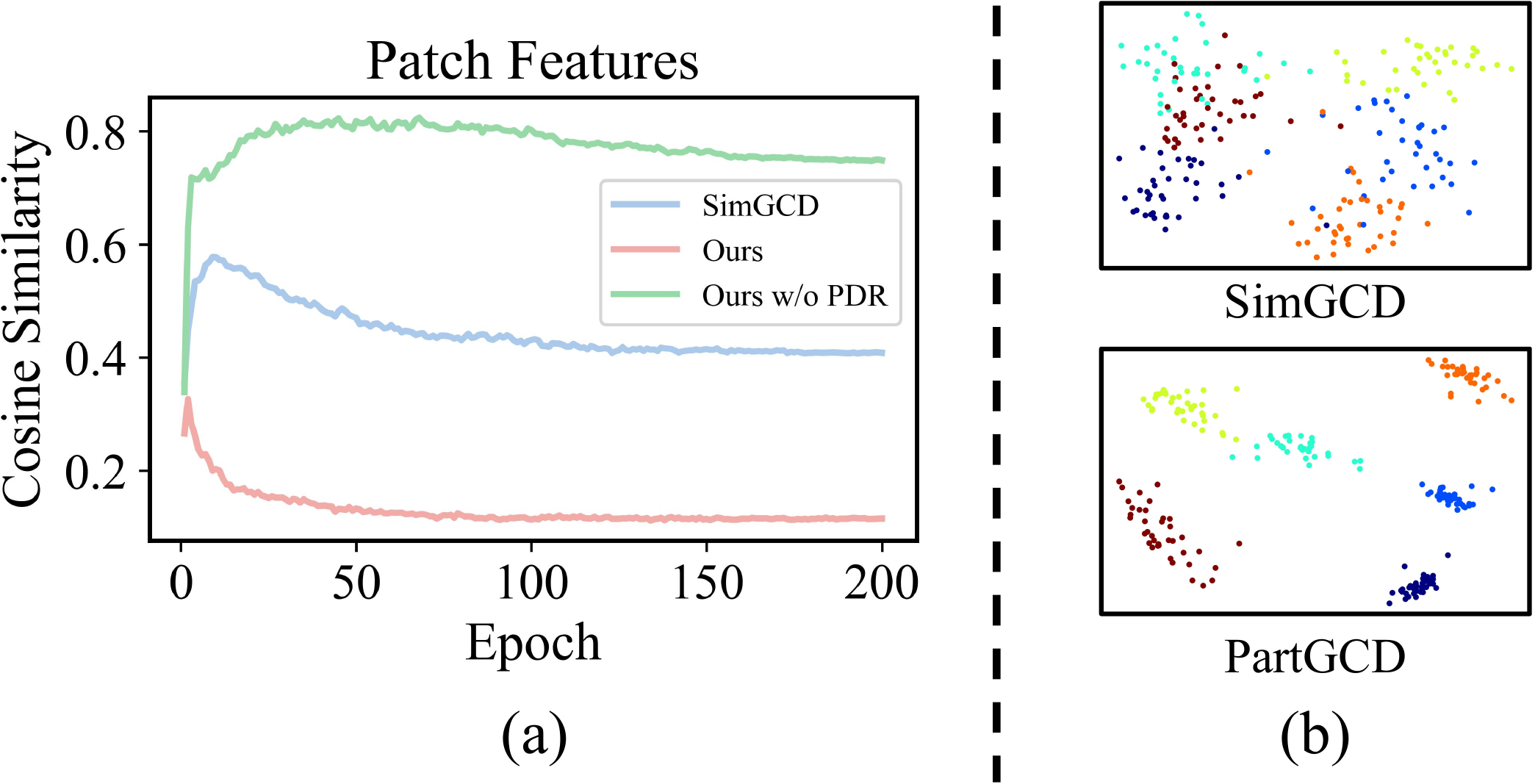 }
    \vspace{-6.5mm}
    \caption{(a) the mean similarity of patch features for each image averaged across the dataset throughout training. (b) the distribution of part features of a random class
    }
\label{fig:similarity}
    \vspace{-6mm}
\end{figure}

\paragraph{Effectiveness of main components.}
\cref{tab:diff_com} presents the ablation results of different objectives.
Comparing (1) and (2) shows that enhancing semantic differences among parts alone improves the baseline method's performance, as it strengthens the model's local perception. The comparison between (4) and (6) highlights the importance of the adapter, as it can selectively emphasize parts relevant to each category through aggregation. The performance drop on old classes in CUB 
could be the adapter's design and optimization, which considers both new and old classes simultaneously. Training on CUB may have led the adapter to focus more on new classes. However, it shows a 3\% performance gain on All, proving its effectiveness.
The suboptimal result of (3) underscores the importance of contrastive learning on aggregated features, as it boosts the discriminative power of these features. In~\cref{fig:similarity} (a), we present the similarity statistics of patch features on the SCars dataset, demonstrating our PDR enhances patch feature diversity. 
\cref{fig:similarity} (b) proves our PartGCD encourages clearer semantic separation among part features.

\paragraph{Effectiveness of our candidate selection method.}
\cref{tab:diff_can} presents experimental results on SCars and CUB using different candidate selection methods. (1) indicates using model predictions to select samples, \ie, $\mathcal{A}_c$, it introduces significant noise to capture object parts. (2) is conducting the Sinkhorn algorithm on the model prediction, showing improvement. 
Our method combines Sinkhorn-adjusted predictions with instance representations, yielding more reliable calibrated prototypes (denoted as CP), which prove to be more effective in (4). 
\begin{table}[t]
    \centering
    \setlength{\tabcolsep}{2.8pt}
  \resizebox{\columnwidth}{!}{
  \begin{tabular}{lccccccccc} \toprule

 &  & \multicolumn{4}{c}{SCars} & \multicolumn{4}{c}{CUB} \\
    \cmidrule(rl){3-6} \cmidrule(rl){7-10}  & Method& \multicolumn{1}{c}{Purity} & \multicolumn{1}{c}{All}  & \multicolumn{1}{c}{Old}  & \multicolumn{1}{c}{New}& \multicolumn{1}{c}{Purity}& \multicolumn{1}{c}{All}  & \multicolumn{1}{c}{Old}  & \multicolumn{1}{c}{New} \\
    \midrule
 (1) & Pred & 56.3& 62.1& 76.0& 55.2& 63.6&  65.8 & 65.9 & 65.8 \\
 (2) & Pred w/ SK  & 62.2&64.5& \textbf{79.9}& 57.0 & 65.6&67.4 &68.3 &67.0    \\
 (3) &CP w/o SK &60.0 &64.1&78.2& 57.3 & 68.6&67.9 & 68.3 &67.7\\
 \rowcolor{mygray}
  (4) &    
 CP
&\textbf{64.9} &\textbf{65.6} &79.5& \textbf{58.9} & \textbf{72.3} & \textbf{68.6}& \textbf{68.9}& \textbf{68.4}\\

    \bottomrule
\end{tabular}}
    \vspace{-3mm}
    \caption{Results of different candidates selection methods.``Purity" is the cluster accuracy of new class samples selected in the final epoch, while ``SK" represents the use of the Sinkhorn algorithm.}
    \label{tab:diff_can}
    \vspace{-5.5mm}
\end{table}

\begin{figure}[!h]
    \centering
    \vspace{-3mm}
    \includegraphics[width=\columnwidth]{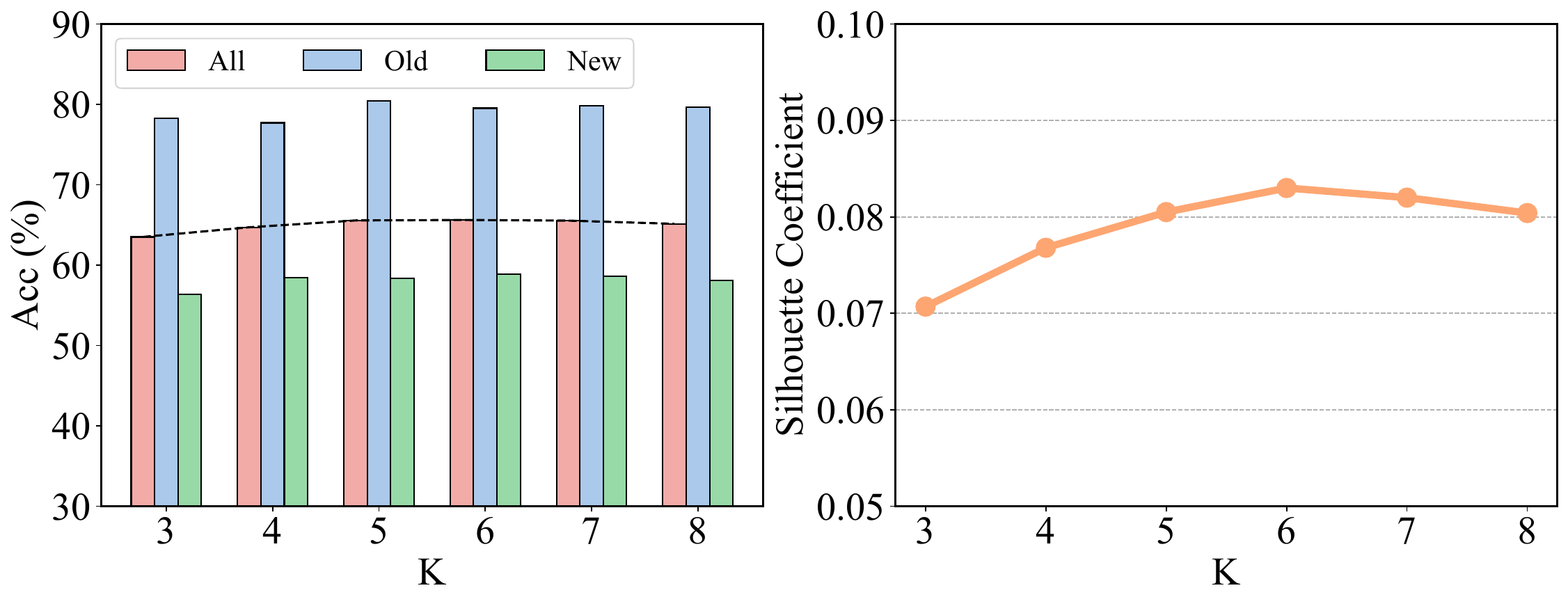 }
    \vspace{-6mm}
    \caption{The sensitivity to $K$ values (left) and the silhouette coefficient (right) on the SCars dataset.
    }
    \label{fig:k_sen}
    \vspace{-5mm}
\end{figure}

\paragraph{The sensitivity to $K$ values.}
In our method, we automatically select the number of parts, \ie, $K$, based on the average silhouette coefficient of $\mathcal{P}_c$. \cref{fig:k_sen} presents a sensitivity analysis of $K$ on the SCars dataset, the experimental results for different $K$ values generally align with changes in the silhouette coefficient, with the optimal performance achieved at $K=6$.
When $K$ is set to 5, 6, or 7, the results differ only slightly, indicating that our method is relatively insensitive to $K$ within a range.

\begin{figure}[!h]
    \centering
    \vspace{-1mm}
    \includegraphics[width=\columnwidth]{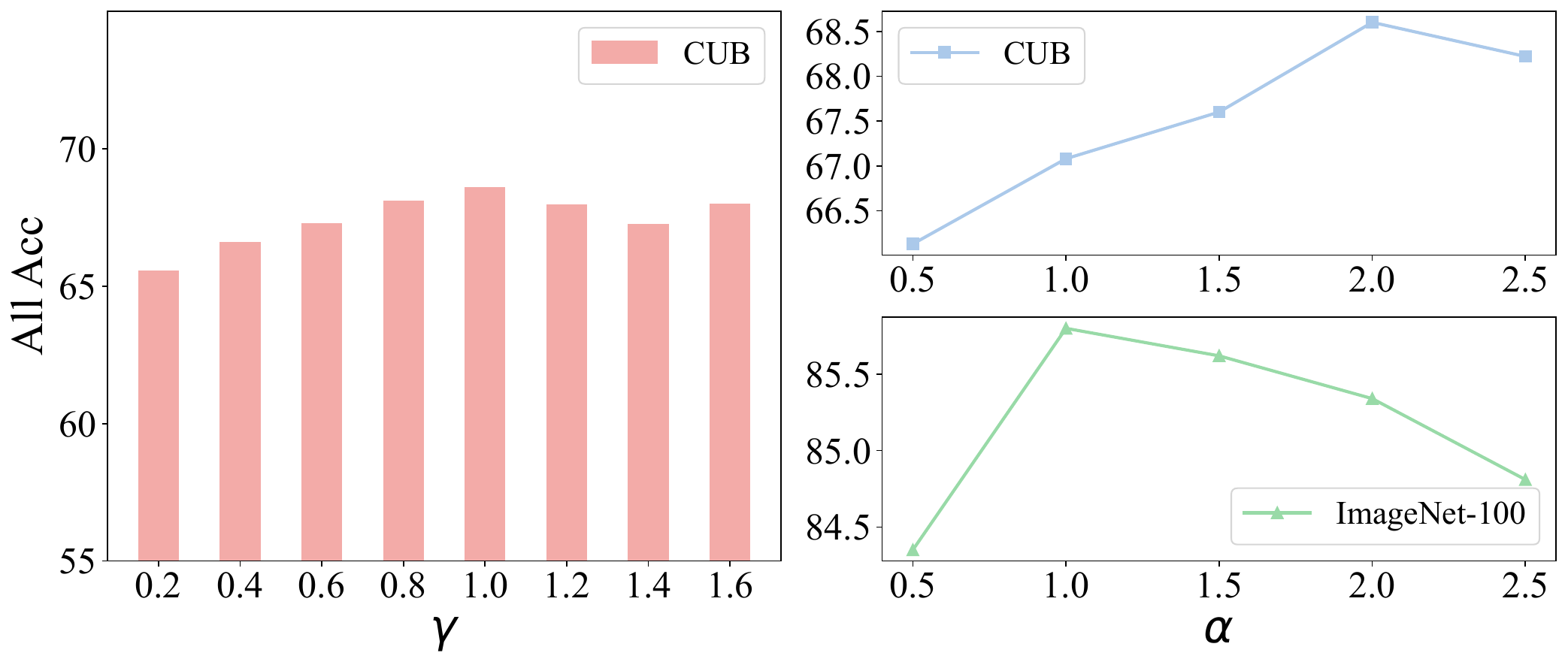 }
    \vspace{-6mm}
    \caption{Impact of hyper-parameters.
    }
    \label{fig:hyperparam_effect}
    \vspace{-4mm}
\end{figure}

\paragraph{Hyperparameters analysis.}
We present the impact of hyperparameters in \cref{fig:hyperparam_effect}. The parameter $\gamma$ controls the number of candidate samples per class, while $\alpha$ balances the importance between global and part features. For fine-grained datasets, higher values of $\alpha$ are required to increase focus on parts. Setting  $\gamma$ too low affects adequate modeling in the GMM, while excessively high values introduce noise. To avoid over-tuning, we set $\gamma=0.8$ for all datasets, $\alpha$ is set to 2 for fine-grained while set to 1 for generic datasets.

\paragraph{The impact of part extraction strategies.}
In \cref{tab:diff_p}, using a prototype method refers to clustering the patch features of $\mathcal{P}_c$ to obtain part prototypes, which then serve as a basis for constructing the attention map by measuring the similarity between each patch feature and prototype. However,
this approach is sensitive to noisy samples and semantic noise in the cluster centers, preventing the accurate representation of a part
and making it less effective than our GMM-based method. The attention-based filtering component functions to remove most background patches, preventing overfitting to background regions, showing to be effective.

\begin{table}[tp]
\small
\centering
\setlength{\tabcolsep}{3pt}
\resizebox{\columnwidth}{!}{
  \begin{tabular}{lccccccccc} \toprule &
 \multirow{2}{*}{GMM}&  \multirow{2}{*}{Prototype} & \multirow{2}{*}{Filter} & \multicolumn{3}{c}{SCars} & \multicolumn{3}{c}{CUB}  \\ 
    \cmidrule(rl){5-7} \cmidrule(rl){8-10} &&&
     & \multicolumn{1}{c}{All} & \multicolumn{1}{c}{Old}  & New & \multicolumn{1}{c}{All} & \multicolumn{1}{c}{Old}  & New  \\  \midrule
     \rowcolor{mygray}
     (1) & \cmark & \xmark & \cmark &\textbf{65.6 }&\textbf{79.5} &\textbf{58.9} & \textbf{68.6}& 68.9 &\textbf{68.4}\\
     (2) & \xmark & \cmark & \cmark &61.7 &77.9& 53.9 & 66.8 & \textbf{72.1}& 64.1 \\
     (3) & \cmark & \xmark & \xmark & 59.8 &75.9 &52.1 &66.1 &71.5 &63.4\\

    \bottomrule
    \end{tabular}}
\vspace{-3mm}
\caption{The impact of part extraction strategies.}
\vspace{-7mm}
\label{tab:diff_p}
\end{table}

\begin{figure*}[t]
    \centering
    \includegraphics[width=\textwidth]{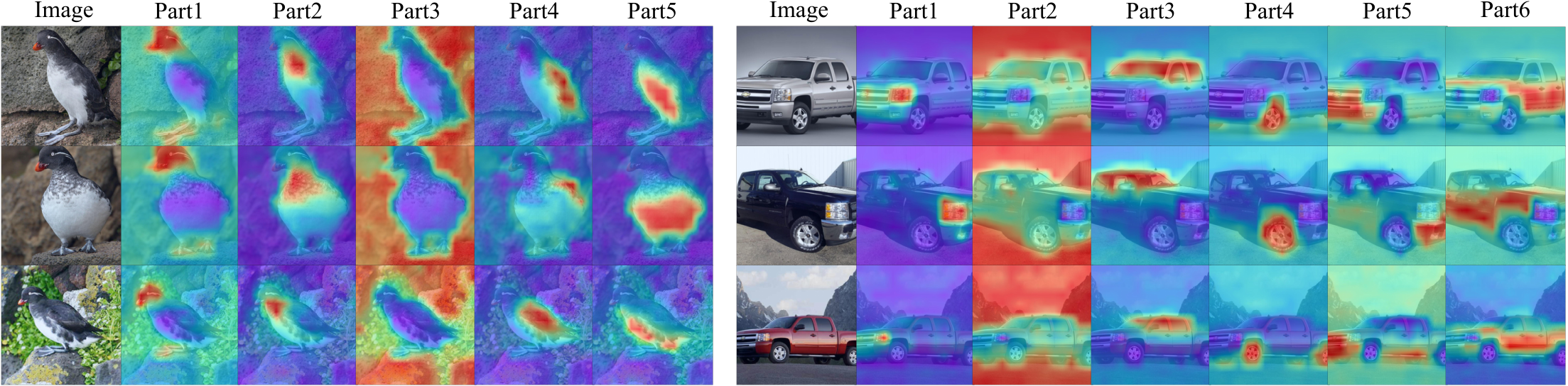 }
    \vspace{-6mm}
    \caption{The part attention maps for three images in the same class for the CUB (left) and SCars (right) datasets.
    }
    \label{fig:pam}
    \vspace{-6mm}
\end{figure*}

\subsection{Qualitative results}
\label{qr}
\paragraph{Part Attention Maps Visualization.}
In~\cref{fig:pam},  we present part attention maps for three images in the same class for the SCars and CUB datasets. The qualitative results show that images within the same class yield semantically consistent responses to the same Gaussian component of their corresponding GMM, representing the same part, while distinct attention maps highlight different semantic parts. We show more visualizations in \textit{supp}.

\begin{figure}[!th]
    \centering
    \includegraphics[width=\columnwidth]{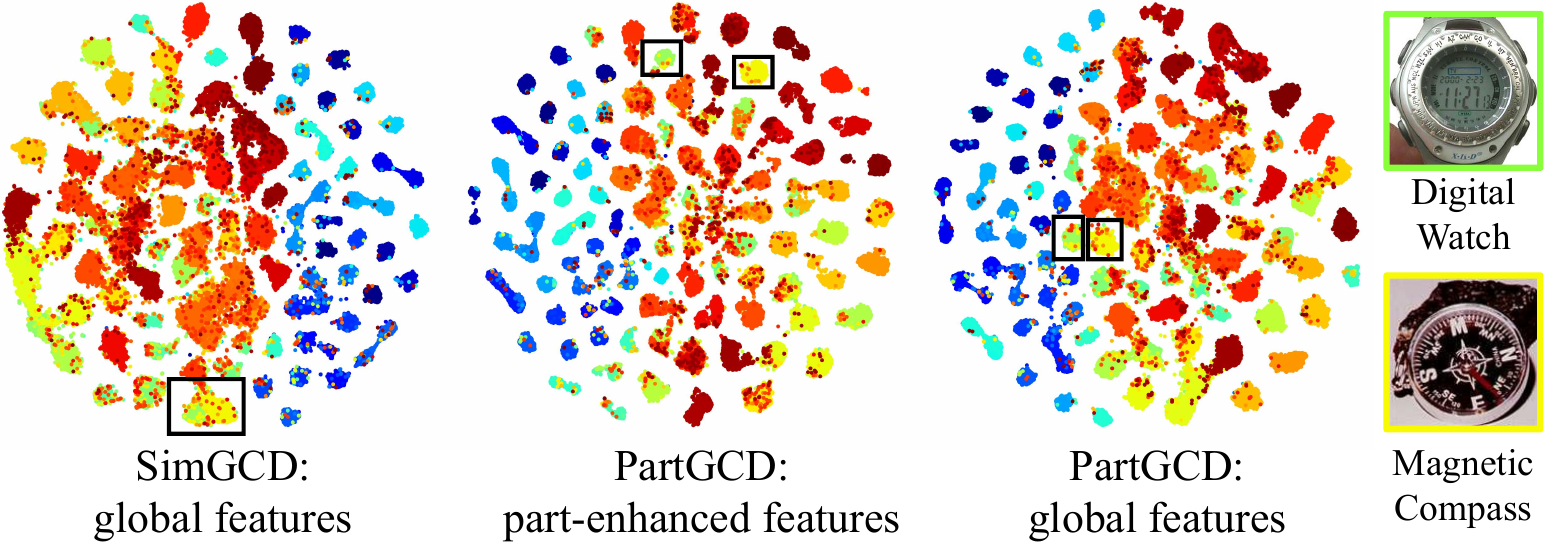}
    \\
    \makebox[0.5\columnwidth]{\small (a) ImageNet-100} 
    \\ \vspace{0.5em}
    \includegraphics[width=\columnwidth]{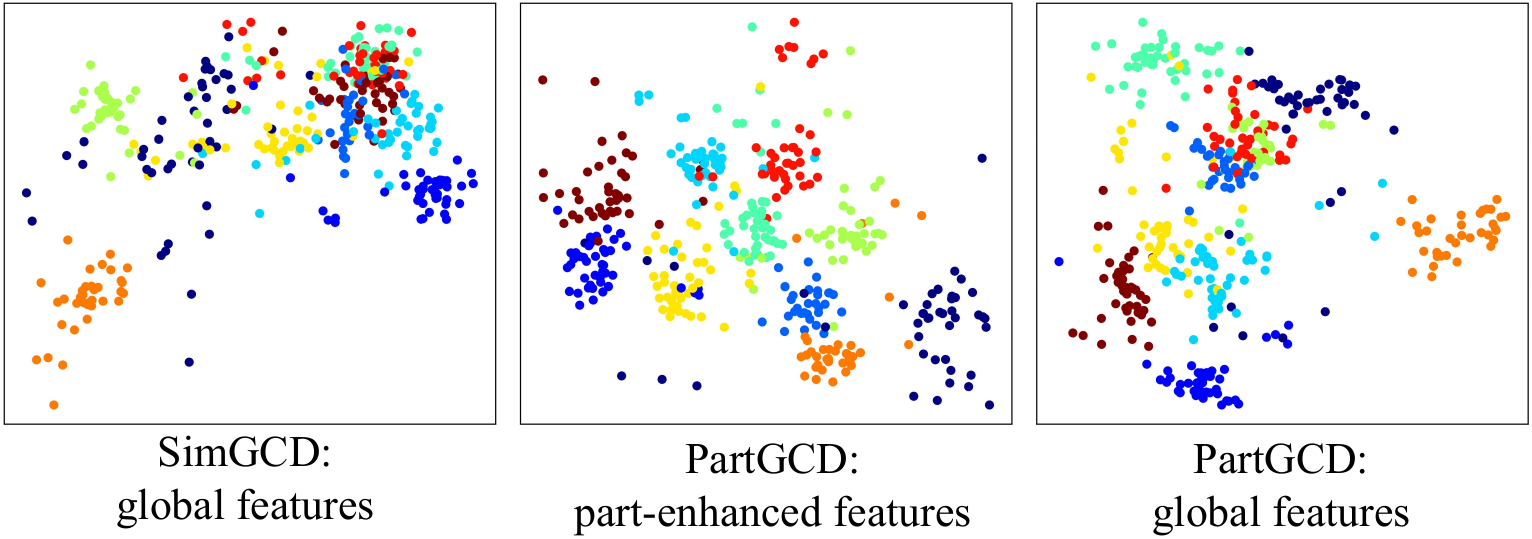}
     \\
    \makebox[0.5\columnwidth]{\small (b) SCars (10 random classes)} 
    \vspace{-2mm}
    \caption{The data distribution of unlabelled samples.}
    \label{fig:datadis}
    \vspace{-4mm}
\end{figure}

\paragraph{Feature Distribution Visualization.}
As shown in \cref{fig:datadis}, we present the distribution of unlabeled samples on the ImageNet-100 and SCars datasets. For ImageNet-100, our method effectively distinguishes visually similar categories, such as ``digital watch” and ``magnetic compass.” The classification accuracy for “digital watch” using our method reaches 87.1\%, vastly surpassing SimGCD's 0.3\%. More performance comparisons on visually similar classes in \textit{supp} further emphasize the effectiveness of PartGCD on generic datasets.
Moreover, our method not only improves the clarity of class boundaries but also forms more compact clusters across both fine-grained and generic datasets, demonstrating that our approach establishes more accurate and robust visual-semantic relationships. Our final classification combines part-enhanced and global features, complementing each other to achieve improved results.

\section{Conclusions
}

In this paper, we propose PartGCD, a novel approach that introduces part knowledge to tackle Gernealized Category Discovery task, especially for challenging fine-grained scenarios.
We first highlight that current methods relying on global features face challenges in fine-grained category discovery, reflecting limitations in the model’s category understanding. 
To address this issue, we introduce an Adaptive Part Decomposition that  extract part features.
 By enhancing the semantic distance between different part features and incorporating them into learning, we facilitate category understanding of the GCD model, establishing more accurate and robust visual-semantic relationships and thus achieving significant improvements.
We hope our research inspires researchers to rethink the importance of local knowledge in category discovery and provides a new direction for GCD.

\paragraph{Limitations and future work}.
Although our method automatically chooses the number of parts for each dataset, a more refined mechanism that selects different number of parts for each category
could still enhance the performance by reducing noisy parts. 
We provide some attempts to address this in the \textit{supp}, aiming to encourage future research.


{
    \small
    \bibliographystyle{ieeenat_fullname}
    \bibliography{main}

\begin{thebibliography}{70}
\providecommand{\natexlab}[1]{#1}
\providecommand{\url}[1]{\texttt{#1}}
\expandafter\ifx\csname urlstyle\endcsname\relax
  \providecommand{\doi}[1]{doi: #1}\else
  \providecommand{\doi}{doi: \begingroup \urlstyle{rm}\Url}\fi

\bibitem[Amir et~al.(2022)Amir, Gandelsman, Bagon, and Dekel]{amir2021deep}
Shir Amir, Yossi Gandelsman, Shai Bagon, and Tali Dekel.
\newblock Deep vit features as dense visual descriptors.
\newblock \emph{ECCVW What is Motion For?}, 2022.

\bibitem[Asano et~al.(2020)Asano, Rupprecht, and Vedaldi]{asano2020self}
Yuki~M. Asano, Christian Rupprecht, and Andrea Vedaldi.
\newblock Self-labelling via simultaneous clustering and representation learning.
\newblock In \emph{International Conference on Learning Representations (ICLR)}, 2020.

\bibitem[Assran et~al.(2022)Assran, Caron, Misra, Bojanowski, Bordes, Vincent, Joulin, Rabbat, and Ballas]{assran2022masked}
Mahmoud Assran, Mathilde Caron, Ishan Misra, Piotr Bojanowski, Florian Bordes, Pascal Vincent, Armand Joulin, Mike Rabbat, and Nicolas Ballas.
\newblock Masked siamese networks for label-efficient learning.
\newblock In \emph{European Conference on Computer Vision}, pages 456--473. Springer, 2022.

\bibitem[Cao et~al.(2022)Cao, Brbic, and Leskovec]{orca}
Kaidi Cao, Maria Brbic, and Jure Leskovec.
\newblock Open-world semi-supervised learning.
\newblock In \emph{International Conference on Learning Representations}, 2022.

\bibitem[Cao et~al.(2024)Cao, Zheng, Wang, Yu, Shen, Li, Lu, and Tian]{legogcd}
Xinzi Cao, Xiawu Zheng, Guanhong Wang, Weijiang Yu, Yunhang Shen, Ke Li, Yutong Lu, and Yonghong Tian.
\newblock Solving the catastrophic forgetting problem in generalized category discovery.
\newblock In \emph{Proceedings of the IEEE/CVF Conference on Computer Vision and Pattern Recognition (CVPR)}, pages 16880--16889, 2024.

\bibitem[Caron et~al.(2021)Caron, Touvron, Misra, J{\'e}gou, Mairal, Bojanowski, and Joulin]{dino}
Mathilde Caron, Hugo Touvron, Ishan Misra, Herv{\'e} J{\'e}gou, Julien Mairal, Piotr Bojanowski, and Armand Joulin.
\newblock Emerging properties in self-supervised vision transformers.
\newblock In \emph{Proceedings of the IEEE/CVF international conference on computer vision}, pages 9650--9660, 2021.

\bibitem[Chen et~al.(2019)Chen, Li, Tao, Barnett, Rudin, and Su]{chen2019looks}
Chaofan Chen, Oscar Li, Daniel Tao, Alina Barnett, Cynthia Rudin, and Jonathan~K Su.
\newblock This looks like that: deep learning for interpretable image recognition.
\newblock \emph{Advances in neural information processing systems}, 32, 2019.

\bibitem[Choi et~al.(2024)Choi, Kang, and Cho]{cms}
Sua Choi, Dahyun Kang, and Minsu Cho.
\newblock Contrastive mean-shift learning for generalized category discovery.
\newblock In \emph{Proceedings of the IEEE/CVF Conference on Computer Vision and Pattern Recognition}, 2024.

\bibitem[Cole et~al.(2022)Cole, Yang, Wilber, Mac~Aodha, and Belongie]{cole2022does}
Elijah Cole, Xuan Yang, Kimberly Wilber, Oisin Mac~Aodha, and Serge Belongie.
\newblock When does contrastive visual representation learning work?
\newblock In \emph{Proceedings of the IEEE/CVF Conference on Computer Vision and Pattern Recognition}, pages 14755--14764, 2022.

\bibitem[Cuturi(2013)]{sinkhorn}
Marco Cuturi.
\newblock Sinkhorn distances: Lightspeed computation of optimal transport.
\newblock \emph{Advances in neural information processing systems}, 26, 2013.

\bibitem[Deng et~al.(2009)Deng, Dong, Socher, Li, Li, and Fei-Fei]{imagnet}
Jia Deng, Wei Dong, Richard Socher, Li-Jua Li, Kai Li, and Li Fei-Fei.
\newblock Imagenet: A large-scale hierarchical image database.
\newblock In \emph{CVPR}, 2009.

\bibitem[Donnelly et~al.(2022)Donnelly, Barnett, and Chen]{donnelly2022deformable}
Jon Donnelly, Alina~Jade Barnett, and Chaofan Chen.
\newblock Deformable protopnet: An interpretable image classifier using deformable prototypes.
\newblock In \emph{Proceedings of the IEEE/CVF conference on computer vision and pattern recognition}, pages 10265--10275, 2022.

\bibitem[Edwards et~al.(2019)Edwards, Williams, Gentner, and Lombrozo]{EDWARDS201921}
Brian~J. Edwards, Joseph~J. Williams, Dedre Gentner, and Tania Lombrozo.
\newblock Explanation recruits comparison in a category-learning task.
\newblock \emph{Cognition}, 185:\penalty0 21--38, 2019.

\bibitem[Fei et~al.(2022)Fei, Zhao, Yang, and Zhao]{xcon}
Yixin Fei, Zhongkai Zhao, Siwei Yang, and Bingchen Zhao.
\newblock Xcon: Learning with experts for fine-grained category discovery.
\newblock In \emph{British Machine Vision Conference (BMVC)}, 2022.

\bibitem[Fini et~al.(2021)Fini, Sangineto, Lathuiliere, Zhong, Nabi, and Ricci]{uno}
Enrico Fini, Enver Sangineto, St{\'e}phane Lathuiliere, Zhun Zhong, Moin Nabi, and Elisa Ricci.
\newblock A unified objective for novel class discovery.
\newblock In \emph{Proceedings of the IEEE/CVF International Conference on Computer Vision}, pages 9284--9292, 2021.

\bibitem[Gu et~al.(2023)Gu, Zhang, Xu, and He]{gu2023class}
P. Gu, C. Zhang, R. Xu, and X. He.
\newblock Class-relation knowledge distillation for novel class discovery.
\newblock In \emph{2023 IEEE/CVF International Conference on Computer Vision (ICCV)}, pages 16428--16437, 2023.

\bibitem[Han et~al.(2019)Han, Vedaldi, and Zisserman]{han2019learning}
Kai Han, Andrea Vedaldi, and Andrew Zisserman.
\newblock Learning to discover novel visual categories via deep transfer clustering.
\newblock In \emph{Proceedings of the IEEE/CVF International Conference on Computer Vision}, pages 8401--8409, 2019.

\bibitem[Han et~al.(2020)Han, Rebuffi, Ehrhardt, Vedaldi, and Zisserman]{rs}
Kai Han, Sylvestre-Alvise Rebuffi, Sebastien Ehrhardt, Andrea Vedaldi, and Andrew Zisserman.
\newblock Automatically discovering and learning new visual categories with ranking statistics.
\newblock In \emph{International Conference on Learning Representations (ICLR)}, 2020.

\bibitem[He et~al.(2022)He, Chen, Liu, Kortylewski, Yang, Bai, and Wang]{he2022transfg}
Ju He, Jie-Neng Chen, Shuai Liu, Adam Kortylewski, Cheng Yang, Yutong Bai, and Changhu Wang.
\newblock Transfg: A transformer architecture for fine-grained recognition.
\newblock In \emph{Proceedings of the AAAI conference on artificial intelligence}, pages 852--860, 2022.

\bibitem[He et~al.(2016)He, Zhang, Ren, and Sun]{he2016deep}
Kaiming He, Xiangyu Zhang, Shaoqing Ren, and Jian Sun.
\newblock Deep residual learning for image recognition.
\newblock In \emph{Proceedings of the IEEE conference on computer vision and pattern recognition}, pages 770--778, 2016.

\bibitem[Hsu et~al.(2018)Hsu, Lv, and Kira]{hsu2017learning}
Yen-Chang Hsu, Zhaoyang Lv, and Zsolt Kira.
\newblock Learning to cluster in order to transfer across domains and tasks.
\newblock In \emph{International Conference on Learning Representations (ICLR)}, 2018.

\bibitem[Hsu et~al.(2019)Hsu, Lv, Schlosser, Odom, and Kira]{hsu2019multi}
Yen-Chang Hsu, Zhaoyang Lv, Joel Schlosser, Phillip Odom, and Zsolt Kira.
\newblock Multi-class classification without multi-class labels.
\newblock In \emph{International Conference on Learning Representations (ICLR)}, 2019.

\bibitem[Huang et~al.(2016)Huang, Xu, Tao, and Zhang]{huang2016part}
Shaoli Huang, Zhe Xu, Dacheng Tao, and Ya Zhang.
\newblock Part-stacked cnn for fine-grained visual categorization.
\newblock In \emph{Proceedings of the IEEE conference on computer vision and pattern recognition}, pages 1173--1182, 2016.

\bibitem[Huang and Li(2020)]{huang2020interpretable}
Zixuan Huang and Yin Li.
\newblock Interpretable and accurate fine-grained recognition via region grouping.
\newblock In \emph{Proceedings of the IEEE/CVF Conference on Computer Vision and Pattern Recognition}, pages 8662--8672, 2020.

\bibitem[Knight(2008)]{knight2008sinkhorn}
Philip~A Knight.
\newblock The sinkhorn--knopp algorithm: convergence and applications.
\newblock \emph{SIAM Journal on Matrix Analysis and Applications}, 30\penalty0 (1):\penalty0 261--275, 2008.

\bibitem[Krause et~al.(2013)Krause, Stark, Deng, and Fei-Fei]{StanfordCars}
Jonathan Krause, Michael Stark, Jia Deng, and Li Fei-Fei.
\newblock 3d object representations for fine-grained categorization.
\newblock In \emph{4th International IEEE Workshop on 3D Representation and Recognition (3dRR-13)}, 2013.

\bibitem[Krizhevsky and Hinton(2009)]{cifar}
Alex Krizhevsky and Geoffrey Hinton.
\newblock Learning multiple layers of features from tiny images.
\newblock \emph{Technical report}, 2009.

\bibitem[Krizhevsky et~al.(2017)Krizhevsky, Sutskever, and Hinton]{krizhevsky2017imagenet}
Alex Krizhevsky, Ilya Sutskever, and Geoffrey~E Hinton.
\newblock Imagenet classification with deep convolutional neural networks.
\newblock \emph{Communications of the ACM}, 60\penalty0 (6):\penalty0 84--90, 2017.

\bibitem[Kuhn(1955)]{hungarian}
Harold~W Kuhn.
\newblock The hungarian method for the assignment problem.
\newblock \emph{Naval research logistics quarterly}, 1955.

\bibitem[Li et~al.(2023)Li, Fan, Huo, and Gao]{li2023modeling}
Wenbin Li, Zhichen Fan, Jing Huo, and Yang Gao.
\newblock Modeling inter-class and intra-class constraints in novel class discovery.
\newblock In \emph{Proceedings of the IEEE/CVF Conference on Computer Vision and Pattern Recognition}, pages 3449--3458, 2023.

\bibitem[Liang et~al.(2022)Liang, Wang, Miao, and Yang]{liang2022gmmseg}
Chen Liang, Wenguan Wang, Jiaxu Miao, and Yi Yang.
\newblock Gmmseg: Gaussian mixture based generative semantic segmentation models.
\newblock \emph{Advances in Neural Information Processing Systems}, 35:\penalty0 31360--31375, 2022.

\bibitem[Lin et~al.(2015)Lin, Shen, Lu, and Jia]{lin2015deep}
Di Lin, Xiaoyong Shen, Cewu Lu, and Jiaya Jia.
\newblock Deep lac: Deep localization, alignment and classification for fine-grained recognition.
\newblock In \emph{Proceedings of the IEEE conference on computer vision and pattern recognition}, pages 1666--1674, 2015.

\bibitem[Liu et~al.(2024)Liu, Cai, Jia, Qiu, Wang, and Pu]{Liu_2024_CVPR}
Yu Liu, Yaqi Cai, Qi Jia, Binglin Qiu, Weimin Wang, and Nan Pu.
\newblock Novel class discovery for ultra-fine-grained visual categorization.
\newblock In \emph{Proceedings of the IEEE/CVF Conference on Computer Vision and Pattern Recognition (CVPR)}, pages 17679--17688, 2024.

\bibitem[Maji et~al.(2013)Maji, Rahtu, Kannala, Blaschko, and Vedaldi]{aircraft}
Subhransu Maji, Esa Rahtu, Juho Kannala, Matthew Blaschko, and Andrea Vedaldi.
\newblock Fine-grained visual classification of aircraft.
\newblock \emph{arXiv preprint arXiv:1306.5151}, 2013.

\bibitem[Oquab et~al.(2023)Oquab, Darcet, Moutakanni, Vo, Szafraniec, Khalidov, Fernandez, Haziza, Massa, El-Nouby, et~al.]{oquab2023dinov2}
Maxime Oquab, Timoth{\'e}e Darcet, Th{\'e}o Moutakanni, Huy Vo, Marc Szafraniec, Vasil Khalidov, Pierre Fernandez, Daniel Haziza, Francisco Massa, Alaaeldin El-Nouby, et~al.
\newblock Dinov2: Learning robust visual features without supervision.
\newblock \emph{arXiv preprint arXiv:2304.07193}, 2023.

\bibitem[Park et~al.(2023)Park, Kim, Heo, Kim, and Yun]{park2023ssl}
Namuk Park, Wonjae Kim, Byeongho Heo, Taekyung Kim, and Sangdoo Yun.
\newblock What do self-supervised vision transformers learn?
\newblock In \emph{International Conference on Learning Representations}, 2023.

\bibitem[Parkhi et~al.(2012)Parkhi, Vedaldi, Zisserman, and Jawahar]{parkhi2012cats}
Omkar~M Parkhi, Andrea Vedaldi, Andrew Zisserman, and CV Jawahar.
\newblock Cats and dogs.
\newblock In \emph{2012 IEEE conference on computer vision and pattern recognition}, pages 3498--3505. IEEE, 2012.

\bibitem[Paul et~al.(2024)Paul, Chowdhury, Xiong, Chang, Carlyn, Stevens, Provost, Karpatne, Carstens, Rubenstein, Stewart, Berger-Wolf, Su, and Chao]{paul2024simple}
Dipanjyoti Paul, Arpita Chowdhury, Xinqi Xiong, Feng-Ju Chang, David Carlyn, Samuel Stevens, Kaiya Provost, Anuj Karpatne, Bryan Carstens, Daniel Rubenstein, Charles Stewart, Tanya Berger-Wolf, Yu Su, and Wei-Lun Chao.
\newblock A simple interpretable transformer for fine-grained image classification and analysis.
\newblock In \emph{International Conference on Learning Representations}, 2024.

\bibitem[Peng et~al.(2024)Peng, Wang, Liu, and Cheng]{peng2024let}
Zhimao Peng, Enguang Wang, Xialei Liu, and Ming-Ming Cheng.
\newblock Let's start over: retraining with selective samples for generalized category discovery.
\newblock In \emph{Proceedings of the Thirty-Third International Joint Conference on Artificial Intelligence}, pages 4815--4823, 2024.

\bibitem[Pu et~al.(2023)Pu, Zhong, and Sebe]{dccl}
Nan Pu, Zhun Zhong, and Nicu Sebe.
\newblock Dynamic conceptional contrastive learning for generalized category discovery.
\newblock In \emph{Proceedings of the IEEE/CVF Conference on Computer Vision and Pattern Recognition (CVPR)}, pages 7579--7588, 2023.

\bibitem[Rastegar et~al.(2023)Rastegar, Doughty, and Snoek]{infosieve}
Sarah Rastegar, Hazel Doughty, and Cees Snoek.
\newblock Learn to categorize or categorize to learn? self-coding for generalized category discovery.
\newblock In \emph{Thirty-seventh Conference on Neural Information Processing Systems}, 2023.

\bibitem[Rastegar et~al.(2024)Rastegar, Salehi, Asano, Doughty, and Snoek]{selex}
Sarah Rastegar, Mohammadreza Salehi, Yuki~M Asano, Hazel Doughty, and Cees G~M Snoek.
\newblock Selex: Self-expertise in fine-grained generalized category discovery.
\newblock In \emph{European Conference on Computer Vision}, 2024.

\bibitem[Rousseeuw(1987)]{ROUSSEEUW198753}
Peter~J. Rousseeuw.
\newblock Silhouettes: A graphical aid to the interpretation and validation of cluster analysis.
\newblock \emph{Journal of Computational and Applied Mathematics}, 20:\penalty0 53--65, 1987.

\bibitem[Tan et~al.(2019)Tan, Liu, Ambrose, Tulig, and Belongie]{herbarium}
Kiat~Chuan Tan, Yulong Liu, Barbara Ambrose, Melissa Tulig, and Serge Belongie.
\newblock The herbarium challenge 2019 dataset.
\newblock In \emph{Workshop on Fine-Grained Visual Categorization}, 2019.

\bibitem[Vaze et~al.(2022{\natexlab{a}})Vaze, Han, Vedaldi, and Zisserman]{gcd}
Sagar Vaze, Kai Han, Andrea Vedaldi, and Andrew Zisserman.
\newblock Generalized category discovery.
\newblock In \emph{Proceedings of the IEEE/CVF Conference on Computer Vision and Pattern Recognition}, pages 7492--7501, 2022{\natexlab{a}}.

\bibitem[Vaze et~al.(2022{\natexlab{b}})Vaze, Han, Vedaldi, and Zisserman]{vaze2022openset}
Sagar Vaze, Kai Han, Andrea Vedaldi, and Andrew Zisserman.
\newblock Open-set recognition: a good closed-set classifier is all you need?
\newblock In \emph{International Conference on Learning Representations}, 2022{\natexlab{b}}.

\bibitem[Vaze et~al.(2023)Vaze, Vedaldi, and Zisserman]{mugcd}
Sagar Vaze, Andrea Vedaldi, and Andrew Zisserman.
\newblock No representation rules them all in category discovery.
\newblock \emph{Advances in Neural Information Processing Systems 37}, 2023.

\bibitem[Wah et~al.(2011)Wah, Branson, Welinder, Perona, and Belongie]{cub}
Catherine Wah, Steve Branson, Peter Welinder, Pietro Perona, and Serge Belongie.
\newblock {The Caltech-UCSD Birds-200-2011 Dataset}.
\newblock Technical Report CNS-TR-2011-001, California Institute of Technology, 2011.

\bibitem[Wang et~al.(2023{\natexlab{a}})Wang, Chen, Liu, McCarthy, Frazer, and Carneiro]{wang2023mixture}
Chong Wang, Yuanhong Chen, Fengbei Liu, Davis~James McCarthy, Helen Frazer, and Gustavo Carneiro.
\newblock Mixture of gaussian-distributed prototypes with generative modelling for interpretable image classification.
\newblock \emph{arXiv preprint arXiv:2312.00092}, 2023{\natexlab{a}}.

\bibitem[Wang et~al.(2024{\natexlab{a}})Wang, Peng, Xie, Liu, and Cheng]{wang2024get}
Enguang Wang, Zhimao Peng, Zhengyuan Xie, Xialei Liu, and Ming-Ming Cheng.
\newblock Get: Unlocking the multi-modal potential of clip for generalized category discovery.
\newblock \emph{arXiv preprint arXiv:2403.09974}, 2024{\natexlab{a}}.

\bibitem[Wang et~al.(2022)Wang, Xia, Li, Mao, Feng, Chen, and Zhao]{wang2022solar}
Haobo Wang, Mingxuan Xia, Yixuan Li, Yuren Mao, Lei Feng, Gang Chen, and Junbo Zhao.
\newblock Solar: Sinkhorn label refinery for imbalanced partial-label learning.
\newblock \emph{Advances in neural information processing systems}, 35:\penalty0 8104--8117, 2022.

\bibitem[Wang et~al.(2024{\natexlab{b}})Wang, Vaze, and Han]{sptnet}
Hongjun Wang, Sagar Vaze, and Kai Han.
\newblock Sptnet: An efficient alternative framework for generalized category discovery with spatial prompt tuning.
\newblock In \emph{International Conference on Learning Representations (ICLR)}, 2024{\natexlab{b}}.

\bibitem[Wang et~al.(2021)Wang, Liu, Wang, and Jing]{wang2021interpretable}
Jiaqi Wang, Huafeng Liu, Xinyue Wang, and Liping Jing.
\newblock Interpretable image recognition by constructing transparent embedding space.
\newblock In \emph{Proceedings of the IEEE/CVF international conference on computer vision}, pages 895--904, 2021.

\bibitem[Wang et~al.(2023{\natexlab{b}})Wang, Zhong, Qiao, Cheng, Zheng, Liu, Sebe, Ji, and Chen]{tida}
Yu Wang, Zhun Zhong, Pengchong Qiao, Xuxin Cheng, Xiawu Zheng, Chang Liu, Nicu Sebe, Rongrong Ji, and Jie Chen.
\newblock Discover and align taxonomic context priors for open-world semi-supervised learning.
\newblock In \emph{Thirty-seventh Conference on Neural Information Processing Systems}, 2023{\natexlab{b}}.

\bibitem[Wang et~al.(2020)Wang, Wang, Yang, Li, Li, and Li]{wang2020weakly}
Zhihui Wang, Shijie Wang, Shuhui Yang, Haojie Li, Jianjun Li, and Zezhou Li.
\newblock Weakly supervised fine-grained image classification via guassian mixture model oriented discriminative learning.
\newblock In \emph{Proceedings of the IEEE/CVF conference on computer vision and pattern recognition}, pages 9749--9758, 2020.

\bibitem[Wei et~al.(2016)Wei, Xie, and Wu]{wei2016mask}
Xiu-Shen Wei, Chen-Wei Xie, and Jianxin Wu.
\newblock Mask-cnn: Localizing parts and selecting descriptors for fine-grained image recognition.
\newblock \emph{arXiv preprint arXiv:1605.06878}, 2016.

\bibitem[Wei et~al.(2021)Wei, Song, Mac~Aodha, Wu, Peng, Tang, Yang, and Belongie]{wei2021fine}
Xiu-Shen Wei, Yi-Zhe Song, Oisin Mac~Aodha, Jianxin Wu, Yuxin Peng, Jinhui Tang, Jian Yang, and Serge Belongie.
\newblock Fine-grained image analysis with deep learning: A survey.
\newblock \emph{IEEE transactions on pattern analysis and machine intelligence}, 44\penalty0 (12):\penalty0 8927--8948, 2021.

\bibitem[Wen et~al.(2023)Wen, Zhao, and Qi]{simgcd}
Xin Wen, Bingchen Zhao, and Xiaojuan Qi.
\newblock Parametric classification for generalized category discovery: A baseline study.
\newblock In \emph{Proceedings of the IEEE/CVF International Conference on Computer Vision}, pages 16590--16600, 2023.

\bibitem[Williams and Lombrozo(2010)]{10.5555/1854509.1854762}
Joseph Williams and Tania Lombrozo.
\newblock The role of explanation in discovery and generalization: evidence from category learning.
\newblock In \emph{Proceedings of the 9th International Conference of the Learning Sciences - Volume 2}, page 490–491. International Society of the Learning Sciences, 2010.

\bibitem[Xiao et~al.(2015)Xiao, Xu, Yang, Zhang, Peng, and Zhang]{xiao2015application}
Tianjun Xiao, Yichong Xu, Kuiyuan Yang, Jiaxing Zhang, Yuxin Peng, and Zheng Zhang.
\newblock The application of two-level attention models in deep convolutional neural network for fine-grained image classification.
\newblock In \emph{Proceedings of the IEEE conference on computer vision and pattern recognition}, pages 842--850, 2015.

\bibitem[Yang et~al.(2022)Yang, Zhu, Yu, Wu, and Deng]{yang2022divide}
Muli Yang, Yuehua Zhu, Jiaping Yu, Aming Wu, and Cheng Deng.
\newblock Divide and conquer: Compositional experts for generalized novel class discovery.
\newblock In \emph{Proceedings of the IEEE/CVF Conference on Computer Vision and Pattern Recognition}, pages 14268--14277, 2022.

\bibitem[Yu et~al.(2021)Yu, Zhao, Gao, Yuan, and Xiong]{yu2021benchmark}
Xiaohan Yu, Yang Zhao, Yongsheng Gao, Xiaohui Yuan, and Shengwu Xiong.
\newblock Benchmark platform for ultra-fine-grained visual categorization beyond human performance.
\newblock In \emph{Proceedings of the IEEE/CVF International Conference on Computer Vision}, pages 10285--10295, 2021.

\bibitem[Zhang et~al.(2016)Zhang, Xu, Elhoseiny, Huang, Zhang, Elgammal, and Metaxas]{zhang2016spda}
Han Zhang, Tao Xu, Mohamed Elhoseiny, Xiaolei Huang, Shaoting Zhang, Ahmed Elgammal, and Dimitris Metaxas.
\newblock Spda-cnn: Unifying semantic part detection and abstraction for fine-grained recognition.
\newblock In \emph{Proceedings of the IEEE conference on computer vision and pattern recognition}, pages 1143--1152, 2016.

\bibitem[Zhang et~al.(2014)Zhang, Donahue, Girshick, and Darrell]{zhang2014part}
Ning Zhang, Jeff Donahue, Ross Girshick, and Trevor Darrell.
\newblock Part-based r-cnns for fine-grained category detection.
\newblock In \emph{Computer Vision--ECCV 2014: 13th European Conference, Zurich, Switzerland, September 6-12, 2014, Proceedings, Part I 13}, pages 834--849. Springer, 2014.

\bibitem[Zhang et~al.(2023)Zhang, Khan, Shen, Naseer, Chen, and Khan]{promptcal}
Sheng Zhang, Salman Khan, Zhiqiang Shen, Muzammal Naseer, Guangyi Chen, and Fahad~Shahbaz Khan.
\newblock Promptcal: Contrastive affinity learning via auxiliary prompts for generalized novel category discovery.
\newblock In \emph{Proceedings of the IEEE/CVF Conference on Computer Vision and Pattern Recognition}, pages 3479--3488, 2023.

\bibitem[Zhao and Han(2021)]{zhao21novel}
Bingchen Zhao and Kai Han.
\newblock Novel visual category discovery with dual ranking statistics and mutual knowledge distillation.
\newblock In \emph{Conference on Neural Information Processing Systems (NeurIPS)}, 2021.

\bibitem[Zhao et~al.(2023)Zhao, Wen, and Han]{gpc}
Bingchen Zhao, Xin Wen, and Kai Han.
\newblock Learning semi-supervised gaussian mixture models for generalized category discovery.
\newblock In \emph{Proceedings of the IEEE/CVF International Conference on Computer Vision (ICCV)}, pages 16623--16633, 2023.

\bibitem[Zheng et~al.(2024)Zheng, Pu, Li, Sebe, and Zhong]{zheng2024textual}
Haiyang Zheng, Nan Pu, Wenjing Li, Nicu Sebe, and Zhun Zhong.
\newblock Textual knowledge matters: Cross-modality co-teaching for generalized visual class discovery.
\newblock \emph{arXiv preprint arXiv:2403.07369}, 2024.

\bibitem[Zhong et~al.(2021{\natexlab{a}})Zhong, Fini, Roy, Luo, Ricci, and Sebe]{Zhong_2021_CVPR}
Zhun Zhong, Enrico Fini, Subhankar Roy, Zhiming Luo, Elisa Ricci, and Nicu Sebe.
\newblock Neighborhood contrastive learning for novel class discovery.
\newblock In \emph{Proceedings of the IEEE/CVF Conference on Computer Vision and Pattern Recognition}, pages 10867--10875, 2021{\natexlab{a}}.

\bibitem[Zhong et~al.(2021{\natexlab{b}})Zhong, Zhu, Luo, Li, Yang, and Sebe]{zhong2021openmix}
Zhun Zhong, Linchao Zhu, Zhiming Luo, Shaozi Li, Yi Yang, and Nicu Sebe.
\newblock Openmix: Reviving known knowledge for discovering novel visual categories in an open world.
\newblock In \emph{Proceedings of the IEEE/CVF Conference on Computer Vision and Pattern Recognition}, pages 9462--9470, 2021{\natexlab{b}}.

\end{thebibliography}
}
\clearpage
\maketitlesupplementary

This \textit{Supplementary Material} includes the following sections:
\begin{itemize}
    \item The contrastive representation learning (\cref{sec:crl}).
    \item Experimental setup (\cref{sec:es}).
    \item More comparison results (\cref{sec:fg}).

    \item Aditional ablation and analysis (\cref{sec:er}).
        
    \item Pseudo-code of our method (\cref{sec:pc}).
    \item  Theoretical proofs  (\cref{sec:tj}).
    \item More visualizations (\cref{sec:vis}).
\end{itemize}

\section{Contrastive Representation Learning}
\label{sec:crl}
Contrastive representation learning is widely used in GCD methods as it equips models with instance discrimination capabilities, facilitating the establishment of visual-semantic relationships for unlabeled data.

The objectives of contrastive representation learning consist of unsupervised contrastive loss $\mathcal{L}^{u}_\text{rep}(\boldsymbol{f}^{cls})$ over all samples and supervised contrastive loss $\mathcal{L}^{s}_\text{rep}(\boldsymbol{f}^{cls})$ on labeled data. Specifically, given global feature $\boldsymbol{f}_i^{cls}$ of image $\boldsymbol{x}_i$, we first pass it through a projector (MLP) to obtain the projected feature $\boldsymbol{h}_i$. The unsupervised contrastive $\mathcal{L}^{u}_\text{rep}(\boldsymbol{f}^{cls})$ is defined as:
\begin{equation}
\mathcal{L}^{u}_\text{rep}(\boldsymbol{f}^{cls})=-\frac{1}{|B|}\sum_{i \in B} \log \frac{\exp (\frac{\boldsymbol{h}_i^{\top}  \boldsymbol{h}'_i}{\tau_{u}})}{\sum_{n \in {B}}^{ n \neq i} \exp ( \frac{\boldsymbol{h}_i^{\top} \boldsymbol{h}_{n}}{\tau_{u}} )} \,,
\end{equation}
where $\boldsymbol{h}'_i$ is the projected feature of another view $\boldsymbol{x}'_i$.  The supervised contrastive loss $\mathcal{L}^{s}_\text{rep}(\boldsymbol{f}^{cls})$ is:
\begin{equation}
\mathcal{L}^{s}_\text{rep}(\boldsymbol{f}^{cls}) \!=\!-\frac{1}{|B_l|}\!\!\sum_{i \in B_l}\!\! \frac{1}{|\mathcal{P}^+_i|}\!\! \sum_{j \in \mathcal{P}^+_i}\!\! \log \!\frac{\exp (\frac{\boldsymbol{h}_i^{\top}  \boldsymbol{h}_j}{\tau_{c}} )}{\sum_{n \in {B}_l}^{ n \neq i} \!\!\exp (\!\frac{\boldsymbol{h}_i^{\top}  \boldsymbol{h}_n}{\tau_{c}} \! )} \,,
\end{equation}
where $\mathcal{P}^+_i$ indicates the positive samples  of  $\boldsymbol{x}_i$, \ie, samples with the same classes as $\boldsymbol{x}_i$. Thus, the overall contrastive representation learning loss $\mathcal{L}_\text{rep}(\boldsymbol{f}^{cls})$ is formulated as follows:
\begin{equation}
    \mathcal{L}_\text{rep}(\boldsymbol{f}^{cls}) = \lambda\,\mathcal{L}^{s}_\text{rep}(\boldsymbol{f}^{cls}) + (1-\lambda)\,\mathcal{L}^{u}_\text{rep}(\boldsymbol{f}^{cls})
\end{equation}
where $\lambda$ is the balance hyperparameter.

\section{Experimental Setup}
\label{sec:es}
\paragraph{Dataset details.}
In the main paper, we provide experimental results on three generic datasets (CIFAR-10/100~\cite{cifar}, ImageNet-100~\cite{imagnet}), three fine-grained datasets from Semantic Shift Benchmark~\cite{vaze2022openset} (CUB~\cite{cub}, Stanford Cars~\cite{StanfordCars}, FGVC-Aircraft~\cite{aircraft}),  a large scale unbalanced fine-grained dataset (Herbarium19~\cite{herbarium}) and three ultra-fine-grained datasets. For the ultra-fine-grained datasets, RAPL~\cite{Liu_2024_CVPR} introduces five datasets of soybean leaf samples spanning five consecutive growth stages for category discovery tasks.
We use the datasets of its first three stages, \ie, SoyAgeing-\{R1, R3, R4\}, where each class corresponds to the leaves of a distinct soybean cultivar.
Meanwhile, in this supplementary material, we further present results on the Oxford-Pet dataset~\cite{parkhi2012cats} in~\cref{subtab:pet}. 
For all datasets, we follow the standard dataset split protocol~\cite{gcd}: half of the samples from old classes are used as labeled data, while the remaining half, along with all new classes' samples, are treated as unlabeled data.  
\cref{tab:dataset_stats} provides the statistics of the dataset splits.

\begin{table}[h]
      \small
      \vspace{-2mm} 
      \centering
\setlength{\tabcolsep}{2.5pt}      
     \resizebox{\columnwidth}{!}{%
      \begin{tabular}{lcccc}
        \toprule
        \multirow{2}[3]{*}{Dataset}  &
        \multicolumn{2}{c}{Labelled dataset} & \multicolumn{2}{c}{Unlabelled dataset}\\
        \cmidrule(lr){2-3} \cmidrule(lr){4-5}
         & \#Images & \#Classes & \#Images & \#Classes \\
        \midrule
        CIFAR10 & 12.5K & 5 & 37.5K & 10 \\
        CIFAR100 & 20.0K & 80 & 30.0K & 100 \\
        ImageNet-100 & 31.9K & 50 & 95.3K & 100 \\
        \midrule
        CUB & 1.5K & 100 & 4.5K & 200 \\
        Stanford Cars & 2.0K & 98 & 6.1K & 196 \\
        FGVC-Aircraft & 1.7K & 50 & 5.0K & 100 \\
        Oxford-Pet & 942 & 19 & 2738 & 37 \\
        Herbarium 19 & 8.9K & 341 & 25.4K & 683 \\
        \midrule
        SoyAgeing-\{R1, R3, R4\}  & 396 & 99 & 1188 & 198 \\
        \bottomrule
      \end{tabular}}
      \vspace{-2mm}
      
       \caption{The statistics of the dataset splits.}
      \vspace{-5mm}
       
      \label{tab:dataset_stats}
    \end{table}

\paragraph{Evaluation protocol.}
Following previous works, we evaluate the clustering accuracy (ACC)  of the unlabeled dataset: ${ACC}=\frac{1}{|\mathcal{D}_u|}\sum_{i=1}^{|\mathcal{D}_u|}{\mathbbm{1}[{y}_i={Per}(\hat{y}_i)]}$, where ${Per(\cdot)}$ is the optimal permutation to obtain the highest $ACC$ and can be computed with Hungarian algorithm~\cite{hungarian}.

\paragraph{Implementation details.}
We use a DINO~\cite{dino} pretrained ViT-B/16 as the backbone, producing 196 patch features and 1 CLS feature, \ie, $N_p$ = 196. Following previous~\cite{gcd,simgcd}, we only fine-tune the last transformer block with 200 epochs. We also provide results using a DINOv2~\cite{oquab2023dinov2} pretrained ViT-B/14 backbone in~\cref{subtab:dino2}, producing $N_p$ = 257.
According to the analysis in~\cref{ssec:ablations}, $\alpha$ is set to 2 for all fine-grained datasets
and 1 for all generic datasets, while $\gamma$ is set to 1 for all datasets. 
We warm up 20 epochs for fine-grained datasets and 10 epochs for generic datasets, and $\alpha$ is linearly increasing from 0 during warm-up.
 Align with~\cite{gcd}, we only fine-tune the last transformer block for 200 epochs with batch size set to 128 for training and 256 for testing, 
 the learning rate is decayed from 0.1 in a cosine schedule,
 and $\lambda$ is set to 0.35, 
the temperature values $\tau_u$, $\tau_c$ are set to 0.07 and 1.0, respectively.
Following SimGCD~\cite{simgcd}, 
$\tau_s$ is set to 0.1, and $\tau_t$ is initially set to 0.07, then raising to 0.04 with a cosine schedule in the starting 30 epochs.
During the candidate selection stage, we calibrate class prototypes using~\cref{eq5} in main paper. For empty $\mathcal{Q}_c$  in the very early stages of training, we select a number of samples equal to the average count of old-class samples, based on adjusted predictions, to populate $\mathcal{Q}_c$.
We construct the part-based GMM using fixed patch features that before the last trainable block and generate part attention maps based on these fixed features. These part attention maps guide the learnable patch features generated from the final block to form part features.
Due to variations in object pose or random augmentations, not all parts are present in every image. Therefore, our PDR operates only on the parts that appear simultaneously in both views of an image, which can be determined based on the part attention maps.
The GMMs are constructed on GPU.
All experiments are conducted with 1 NVIDIA GeForce RTX3090 GPU.

\begin{figure*}[!t]
    \centering
    \includegraphics[width=1\textwidth]{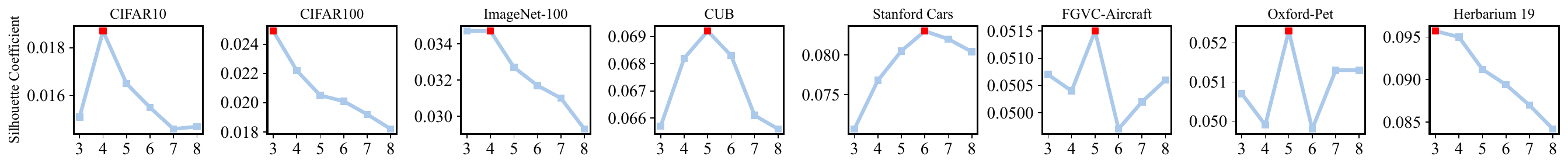}
    \vspace{-5mm}
    \caption{The silhouette coefficients for each dataset, where $x$-axis indicates the  $K$ values from 3 to 8. The selected values are in red color.
    }
    \label{fig:silhouette}
    \vspace{-3mm}
\end{figure*}

\paragraph{Silhouette Coefficient calculation.}
Our method automatically sets the number of parts (\ie, $K$) for the dataset using the silhouette coefficient~\cite{ROUSSEEUW198753}, a classic clustering metric whose higher value indicates that a point is well matched to its cluster, thus eliminating the need for complex parameter tuning.
Specifically, \( K \) is determined based on the average silhouette coefficient computed across the test data of old classes.  
For example, given \( K=5 \), we calculate the silhouette coefficient of the filtered patch features for each old class's test data and then average these values across all old classes.
\cref{fig:silhouette} shows the silhouette coefficients for each dataset with \( K \) ranging from 3 to 8. Based on these results, we select the \( K \) that achieves the highest average silhouette coefficient as the number of parts for each dataset.

\section{Comparisons on Additional Fine-Grained Datasets}
\label{sec:fg}

\paragraph{Comparisons on the Oxford-Pet dataset.}
We present the results of the Oxford-Pet dataset in~\cref{subtab:pet}. Although our method lags behind on ``Old" classes, 
it outperforms the previous best  4\% on ``New"  and slightly leads on ``All". 

\begin{table}[tp]
\small
\centering
\begin{tabular}{lccc}
\toprule
  \multirow{2}{*}{} &   \multicolumn{3}{c}{Oxford-Pet}
\\
\cmidrule(rl){2-4}
  Method    & All  & Old  & New \\
\midrule
GCD~\cite{gcd} &80.2 &85.1& 77.6 \\
DCCL~\cite{dccl} &88.1& 88.2& 88.0 \\
InfoSieve~\cite{infosieve} & 91.8  & \textbf{92.6} & 91.3 \\
SelEx~\cite{selex} & 92.5 & 91.9 & 92.8\\
SimGCD~\cite{simgcd}& 88.8 &82.4 &92.1\\
\midrule
\rowcolor{mygray}
\textbf{PartGCD} & \textbf{92.9} &85.5  & \textbf{96.8} \\
\bottomrule
\end{tabular}
\vspace{-2mm}
\caption{Comparisons on the Oxford-Pet Dataset.}
\label{subtab:pet}
\vspace{-5mm}
\end{table}

\paragraph{Results using DINOv2 backbone}
\cref{subtab:dino2} presents the results using DINOv2~\cite{oquab2023dinov2} as the backbone. While a stronger backbone helps establish visual-semantic relationships for unlabeled datasets, the consistent improvements of our PartGCD across all datasets demonstrate the effectiveness of incorporating part knowledge into learning.

\begin{table}[!ht]
\small
\centering
\vspace{-3mm}
\setlength{\tabcolsep}{1.5pt}
\resizebox{\columnwidth}{!}{%
\begin{tabular}{lccc|ccc|ccc|ccc}
\toprule
  \multirow{2}{*}{} &   \multicolumn{3}{c}{CUB} & \multicolumn{3}{c}{SCars} & \multicolumn{3}{c}{Aircraft}  & \multicolumn{3}{c}{Average}
\\
\cmidrule(rl){2-4}\cmidrule(rl){5-7}\cmidrule(rl){8-10} \cmidrule(rl){11-13}
  Method    & All  & Old  & New  & All  & Old  & New  &  All & Old  & New &   All & Old  & New    \\
\midrule
GCD~\cite{gcd}& 71.9 &71.2& 72.3 &65.7& 67.8 &64.7 &55.4& 47.9 &59.2& 64.3&62.3& 65.4
\\ 
SimGCD~\cite{simgcd} & 71.5& 78.1& 68.3 &71.5 &81.9 &66.6& 63.9& 69.9& 60.9& 69.0&76.6 &65.3 \\
$\mu$GCD~\cite{mugcd} & 74.0& 75.9 &73.1& 76.1 & \textbf{91.0} &68.9 &66.3& 68.7 &65.1& 72.1 & 78.5 & 69.0\\
\midrule
\rowcolor{mygray}
\textbf{PartGCD} & \textbf{77.6} & \textbf{80.6} & \textbf{76.1} & \textbf{78.2} & 88.7& \textbf{73.1} & \textbf{71.1} & \textbf{75.6} & \textbf{68.8} & \textbf{75.6} & \textbf{81.6} & \textbf{72.7} \\
\bottomrule
\end{tabular}}
\vspace{-3mm}

\caption{Results using DINOv2 backbone.}
\label{subtab:dino2}
\vspace{-5mm}
\end{table}

\section{Additional Ablation and Analysis}
\label{sec:er}

\paragraph{Results with the estimated number of classes.}
Following previous works, we assume the number of classes for each dataset is known, and provide experimental results in the main paper. 
In~\cref{tab:est_c}, we provide results using the estimated number of classes for CUB and Stanford Cars datasets, where the number of classes is estimated by the off-the-shelf method from~\cite{xcon,gcd}, \ie, $C = 231$ for CUB and $C = 230$ for Stanford Cars.
Our method's significant performance improvement over baseline approaches demonstrates its effectiveness and robustness in more realistic scenarios.

\begin{table}[!b]
\small
\vspace{-5mm}

    \centering
     \resizebox{\columnwidth}{!}{
\begin{tabular}{lccccccc}
\toprule
\multirow{2}[3]{*}{Method}                                   & \multirow{2}[3]{*}{Known $C$} &\multicolumn{3}{c}{CUB} &\multicolumn{3}{c}{Stanford Cars} \\
 \cmidrule(lr){3-5} \cmidrule(lr){6-8}
& & All  & Old  & New  & All  & Old  & New \\
\midrule
GCD~\cite{gcd}& \ding{51} & 51.3& 56.6& 48.7 &39.0 &57.6& 29.9  \\
SimGCD~\cite{simgcd}      & \ding{51} & 60.3&	{65.6}&	57.7 & 53.8& 71.9 &45.0  \\
$\mu$GCD~\cite{mugcd} & \ding{51} &65.7& 68.0& 64.6& 56.5& 68.1 &50.9\\
\rowcolor{mygray}
\textbf{PartGCD}                     & \ding{51} &
\textbf{68.6}&\textbf{ 68.9} &\textbf{68.4} & \textbf{65.6} & \textbf{79.5}& \textbf{58.9}\\
\midrule
GCD~\cite{gcd}& \ding{55}\,(w/ Est.) & 47.1& 55.1 &44.8 &35.0 &56.0& 24.8 \\
SimGCD~\cite{simgcd}                 & \ding{55}\,(w/ Est.) & 61.5 & 66.4 & 59.1 & 49.1 & 65.1  &41.3\\
$\mu$GCD~\cite{mugcd} & \ding{55}\,(w/ Est.) &
62.0& 60.3 &62.8& 56.3& 66.8& 51.1 \\

\rowcolor{mygray}
\textbf{PartGCD}            & \ding{55}\,(w/ Est.)   & \textbf{66.4} &\textbf{66.5} &\textbf{66.4} &\textbf{64.9} &\textbf{78.7} &\textbf{58.3} \\
\bottomrule
\end{tabular}}
\vspace{-3mm}
\caption{Results with the estimated number of classes}
\label{tab:est_c}
\end{table}

\begin{figure}[ht]
    \centering
    \vspace{-3mm}
\includegraphics[width=0.7\columnwidth]{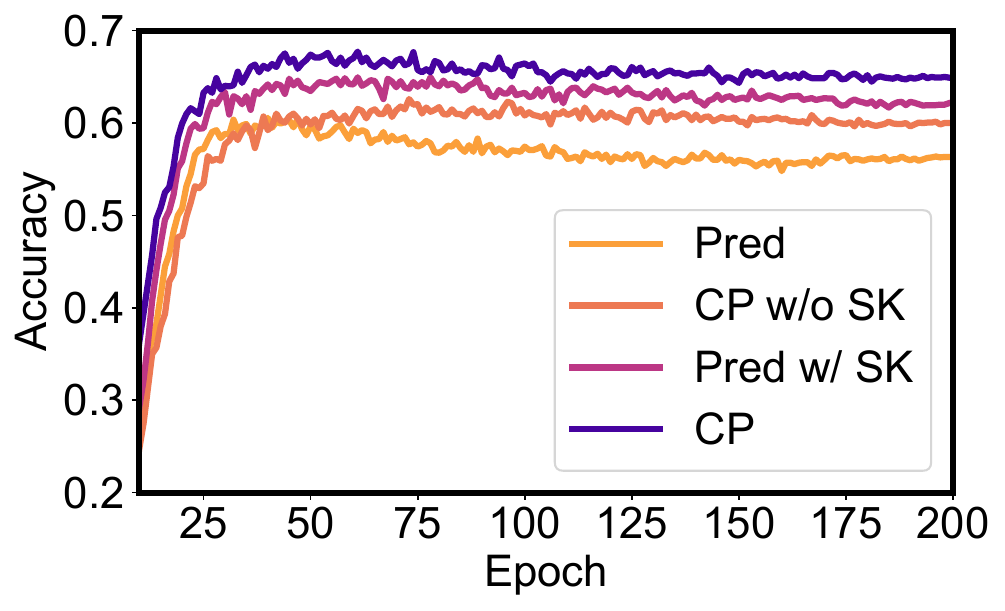}
    \vspace{-3mm}
    \caption{The purity of selected candidates throughout training for CUB datasets with different methods.}
    \label{fig:purity}
    \vspace{-3mm}
\end{figure}

\paragraph{The purity of selected candidates throughout training.}
In the main paper, we evaluate the effectiveness of our candidate selection method in~\cref{ssec:ablations}, and provide results of different candidate selection methods in~\cref{tab:diff_com} in the main paper. Align with~\cref{tab:diff_com}, we show the purity of selected candidates throughout training for
CUB datasets with different methods in~\cref{fig:purity}.

\paragraph{Patch Features to Construct the GMM.}
As mentioned in \textit{Implementation Details},
we construct the part-based GMM using fixed patch features before the last trainable block and generate part attention maps based on these fixed features. We provide results of using different patch features to construct the GMM in~\cref{tab:diff_pa}. 
 Due to the similarity between patch features of each image slightly increases at the beginning of training (see~\cref{fig:similarity} (a) in main paper), 
using learnable patch features from the last block leading to inaccurate GMM fitting and impacting subsequent training. 
Since patch features from the DINO pretrained ViT backbone contain local semantic information and align with object parts~\cite{amir2021deep}, using fixed patch features from the penultimate block shows better results.

\begin{table}[ht]
\vspace{-3mm}
\small
\centering
  \begin{tabular}{ccccc} \toprule
\multirow{2}[3]{*}{Fixed} & \multirow{2}[3]{*}{Learnable} &
\multicolumn{3}{c}{Stanford Cars}  \\ 
     \cmidrule(rl){3-5} 
   & &  \multicolumn{1}{c}{All} & \multicolumn{1}{c}{Old}  & New  \\ \midrule
    \rowcolor{mygray}
    \cmark  & \xmark & \textbf{65.6} &\textbf{79.5} &\textbf{58.9}\\
 \xmark  & \cmark & 63.1 &78.1 &55.8 \\
    \bottomrule
    \end{tabular}
\vspace{-3mm}
    
\caption{Different patch features to construct the GMM.}
\vspace{-5mm}
\label{tab:diff_pa}
\end{table}

\paragraph{Time Efficiency Analysis.}
Following SPTNet~\cite{sptnet}, we provide the performance and time efficiency in~\cref{tab:time},  including total training
epochs, ``All" accuracy, training time per epoch, and inference time.
Compared with SPTNet~\cite{sptnet}, which performs well on generic datasets, our PartGCD significantly outperforms it on the SSB datasets and achieves slightly higher accuracy on the ImageNet-100 dataset. Furthermore, our method is more efficient, requiring only 200 training epochs compared to SPTNet’s 1000 epochs, resulting in fewer total training times.
Compared with SelEx~\cite{selex}, which is designed for fine-grained datasets, our PartGCD achieves higher accuracy on both the SSB and ImageNet-100 datasets, with a notable 2.7\% improvement on ImageNet-100. In terms of time cost, while our total training time on SSB is slightly longer, our inference time is significantly shorter. On a larger-scaled ImageNet-100 dataset, our method demonstrates clear advantages in both training and inference efficiency.
Overall, our method shows better trade-off between performance and efficiency.

\begin{table}[h]
\small
\vspace{-3mm}

    \centering
\setlength{\tabcolsep}{2.5pt}      
     \resizebox{\columnwidth}{!}{
\begin{tabular}{lccccccc}
\toprule
\multirow{2}[3]{*}{Method}     
&\multirow{2}[3]{*}{Epochs}     
&\multicolumn{3}{c}{ImageNet-100} &\multicolumn{3}{c}{SSB}
\\
 \cmidrule(lr){3-5} \cmidrule(lr){6-8} &
& ACC  & Training  & Inference & ACC  & Training  & Inference \\
\midrule
SimGCD~\cite{simgcd} & 200 & 83.0  &160,600s  & 591s  & 56.1 & 12,800s  & 17s \\
SPTNet~\cite{sptnet}  & 1000 &  85.4 & 483,000s &601s& 61.4 &32,000s &17s  \\
SelEX~\cite{selex} & 200 & 83.1& 343,800s & 859s & 63.1&18,000s&61s \\
\rowcolor{mygray}
\textbf{PartGCD}    & 200    & \textbf{85.8}& 329,200s & 630s & \textbf{64.5}&22,800s & 19s               \\
\bottomrule
\end{tabular}}
\vspace{-3mm}
\caption{Time efficiency analysis. We show the total training epochs, ``All" accuracy, total training time (s/epoch $\times$ epochs), and inference time. }
\label{tab:time}
\vspace{-3mm}
\end{table}

\paragraph{Using different $K$ for each category.}
Although our method automatically selects the number of parts (\ie, $K$) for each dataset, introducing a mechanism to determine the optimal number of parts for each individual category could potentially reduce noise and further enhance performance, given the latent differences between categories within the dataset.
However, there are two challenges: lack of new-class samples for $K$ estimation and the need for separate adapters for each class, increasing complexity. 
Nevertheless, as shown in~\cref{tab:diff_k}, we provide two attempts to solve the challenges: 
a) using different $K$ for old classes and applying the average $K$ of old classes to all new classes, and
b) using pseudo-labels generated by off-the-shelf SimGCD for unlabeled data to estimate different $K$ and assign GMMs to samples.  Still, $K$ is estimated using the silhouette coefficient.
The results show that using different $K$ values improves performance. However, it increases costs, such as more adapters and additional trained SimGCD model. We hope this analysis can encourage future work on designing more effective methods.

\begin{table}[ht]
\vspace{-3mm}
\small
\centering
  \begin{tabular}{lccc} \toprule
\multirow{2}[3]{*}{Method}  &
\multicolumn{3}{c}{CUB}  \\ 
     \cmidrule(rl){2-4} &  \multicolumn{1}{c}{All} & \multicolumn{1}{c}{Old}  & New \\
    \midrule 
  Baseline (SimGCD) & 60.3 & 65.6 & 57.7 \\
  Same $K$ for all classes (PartGCD)  & 68.6 & 68.9 & 68.4 \\
  a) different $K$ for old classes &69.6& 71.8& 68.6  \\
  b) different $K$ for all classes &70.2 & 72.4 & 69.1 \\
    \bottomrule
    \end{tabular}
\vspace{-3mm}
\caption{Using different $K$ for each category.}
\vspace{-3mm}
\label{tab:diff_k}
\end{table}

\paragraph{About part attention and part-based approach.}
We focus on the challenges of introducing part knowledge into GCD. Part attention, derived from GMM and Bayes' theorem, explicitly models class-specific parts, offering a more independent and comprehensive approach compared to relying on the model to implicitly learn discriminative cues. 
Additionally, our attention mechanism is equally effective for unlabeled new classes, showing technical innovation, as many fine-grained attention-based methods rely on label-driven approaches.
Beyond the part-based approach, we introduce two other fine-grained approaches into GCD and conduct experimental comparisons in~\cref{tab:beyond_part}.
TransFG~\cite{he2022transfg} uses ViT attention to select discriminative cues, lacking a global perspective and being noise-prone. INTR~\cite{paul2024simple} uses a complex decoder, making optimization difficult and limiting feature learning in GCD. 
In contrast, our part-based method uses all data for GMM modeling, provides a global perspective, can generate refined part features, and uses a simple adapter to emphasize category-specific parts. Our method outperforms alternatives, showing clear advantages.

\begin{table}[!h]
\small
\centering
  \begin{tabular}{lccc} \toprule
\multirow{2}[3]{*}{Method}  &
\multicolumn{3}{c}{CUB}  \\ 
     \cmidrule(rl){2-4} &  \multicolumn{1}{c}{All} & \multicolumn{1}{c}{Old}  & New \\
    \midrule 
  SimGCD & 60.3 & 65.6 & 57.7 \\
 SimGCD + TransFG~\cite{he2022transfg}  &64.4& 69.9& 61.6  \\
 SimGCD + INTR~\cite{paul2024simple}  &64.2& 75.1& 58.7 \\
  PartGCD  & 68.6 & 68.9 & 68.4 \\
 
    \bottomrule
    \end{tabular}
\caption{Using other fine-grained method for GCD.}
\label{tab:beyond_part}
\end{table}

\section{Pseudo-code of our PartGCD}
\label{sec:pc}

The pseudo-code of the proposed PartGCD is presented in~\cref{algo}.

\begin{center}
\begin{algorithm*}[t]
\algsetup{linenosize=\tiny}
\DontPrintSemicolon
\small

  \KwInput{Training dataset $\mathcal{D} = \mathcal{D}_l \cup \mathcal{D}_u$; a backbone to extract global/CLS and patch features; class prototypes serve as cosine classifier; a part adapter to aggregate part features; a projector for contrastive representation learning; $K$ value estimated using the silhouette coefficient.
  }Most frequently mispredicted class
  
  \KwOutput{Class assignments for unlabelled data $\{\hat{y}_i\}_{i=1}^{|\mathcal{D}_u|}$.
  }

    \Repeat{reaching max epochs;}{
    $\{\boldsymbol{\xi}^c\}_{c=1}^C$, $\{a_i\}_{i=1}^{|\mathcal{D}|}$ $\leftarrow$ \cref{eq4} $\sim$ \cref{eq:gmm} \tcp{update GMMs and assignments} \par
      \For{ $(\boldsymbol{x}_i, \boldsymbol{y}_i) \in \text{each batch}$}{
      $p(part_{k}|\boldsymbol{f}^{j}_i)$$\leftarrow$ \cref{eq9} \tcp{the probability that $\boldsymbol{f}^{j}_i$ belongs to $part_k$} \par
      
$\boldsymbol{M}_{i}^{k} = [p({part}_{k}|\boldsymbol{f}^{1}_i);...;p({part}_{k}|\boldsymbol{f}^{N_p}_i) ]$  \tcp{part attention map}\par

$\boldsymbol{v}^{k}_i$ $\leftarrow$ \cref{eq:part_f} \tcp{the $k$-th part feature}\par

$\mathcal{L}_{pdr}$ $\leftarrow$ \cref{eq:loss_pdr} \tcp{Part Discrepancy Regularization}\par
     $\mathcal{L}$ $\leftarrow$ \cref{eq12} and \cref{eq13}  \tcp{final objective }\par
          Optimize the backbone, the class prototypes, the part adapter, and the projector. \par 
      }
  }

  \Return{the final class assignments $\{\hat{y}_i\}_{i=1}^{|\mathcal{D}_u|}$ according to \cref{eq14}
 }

\caption{Pseudo-code of \mbox{{~\OURS}}.}
\label{algo}
\end{algorithm*}
\end{center}

\section{ Theoretical Justification of Sinkhorn}
\label{sec:tj}

In our Adaptive Part Decomposition strategy (\cref{ssec:apd}), we employ the Sinkhorn algorithm to adjust the probability predictions $\boldsymbol{P}$ to a class uniform distribution $\boldsymbol{Q}$, aiming to mitigate the impact of probability bias. In this section, we provide a brief introduction to the derivation of the Sinkhorn-Knopp algorithm to ensure the completeness of our work.

Given a non-negative matrix \( \boldsymbol{P} \in \mathbb{R}^{n \times m}_+ \), the target is to find the closest matrix \( \boldsymbol{Q} \in \mathbb{R}^{n \times m}_+ \) such that each row of 
$\boldsymbol{Q}$ is a probability distribution (\ie, Row-wise constraints) and  each column of $\boldsymbol{Q}$ has equal total mass (\ie, Column-wise constraints). Thus, we can define the optimal transport objective as:
\begin{align}
\begin{split}
     & \min_{\boldsymbol{Q} \geq 0} \sum_{i,j} \boldsymbol{Q}_{ij} \log \frac{\boldsymbol{Q}_{ij}}{\boldsymbol{P}_{ij}}. \\
    s.t. \quad & \text{Row constraints:}   \quad \mathbf{Q} \mathbf{1}_m = \mathbf{1}_n  \\
    & \text{Column constraints:}  \quad \mathbf{Q}^{\top} \mathbf{1}_n = \frac{n}{m} \mathbf{1}_m
\end{split}
\end{align}

By introducing dual variables \( \mathbf{u} \in \mathbb{R}^n \) and \( \mathbf{v} \in \mathbb{R}^m \), we obtain the Lagrangian  formulation of the  optimization problem:
\begin{align}
\begin{split}
    \mathcal{L} = &\sum_{i,j} Q_{ij} \log \frac{Q_{ij}}{P_{ij}} + \sum_{i=1}^n u_i\left(\sum_{j=1}^m Q_{ij} - 1\right) \\
    &+ \sum_{j=1}^m v_j\left(\sum_{i=1}^n Q_{ij} - \frac{n}{m}\right).
\end{split}
\end{align}
Setting \( \frac{\partial \mathcal{L}}{\partial Q_{ij}} = 0 \) yields $ Q_{ij} = P_{ij} e^{-u_i - v_j - 1} = P_{ij} d_i e_j$, where $d_i := e^{-u_i - 0.5}$ and $ e_j := e^{-v_j - 0.5}.$
Let \( \mathbf{D} = \operatorname{diag}(\mathbf{d}) \), \( \mathbf{E} = \operatorname{diag}(\mathbf{e}) \). The optimal \( \mathbf{Q} \) has form: $ \mathbf{Q} = \mathbf{D}\mathbf{P}\mathbf{E}$. 
According to the row constraints and column constraints:
\begin{align}
\begin{split}
     \mathbf{Q} \mathbf{1}_m =  \mathbf{D}\mathbf{P}\mathbf{e} = \mathbf{1}_n,\quad
      \mathbf{Q}^{\top} \mathbf{1}_n = \mathbf{E}\mathbf{P}^{\top}\mathbf{d}=\frac{n}{m} \mathbf{1}_m \,
\end{split}
\end{align}
we can obtain an alternative coordinate descent algorithm for updating row normalization and column normalization:
\begin{equation}
    \mathbf{d} \leftarrow \mathbf{1}_n ./  (\mathbf{P}\mathbf{e}), \quad \mathbf{e} \leftarrow (\frac{n}{m} \mathbf{1}_m) ./ (\mathbf{P}^{\top}\mathbf{d}),
\end{equation}
leading to the alternate between row and column scaling:
\begin{align}
    d_i^{(k)} &= \frac{1}{\sum_{j=1}^m P_{ij} e_j^{(k-1)}} \quad \text{(Row normalization)} \label{eq:row_update} \\
    e_j^{(k)} &= \frac{n/m}{\sum_{i=1}^n d_i^{(k)} P_{ij}} \quad \text{(Column uniformity)} \label{eq:col_update}
\end{align}
which is the Sinkhorn-Knopp iteration~\cite{knight2008sinkhorn}.

The above proof theoretically validates that the Sinkhorn algorithm can achieve our optimization objective. Moreover, this algorithm is widely used in other tasks~\cite{wang2022solar, liang2022gmmseg} and has been empirically verified to be effective in our method (\cref{tab:diff_com} of the main paper). Therefore, we adopt it to adjust prototypes in our Adaptive Part Decomposition, regardless of alternative methods in optimal transport, as they are beyond the scope of this paper.

\section{More Visualizations}
\label{sec:vis}

\begin{figure}[!b]
    \centering
    \includegraphics[width=\columnwidth]{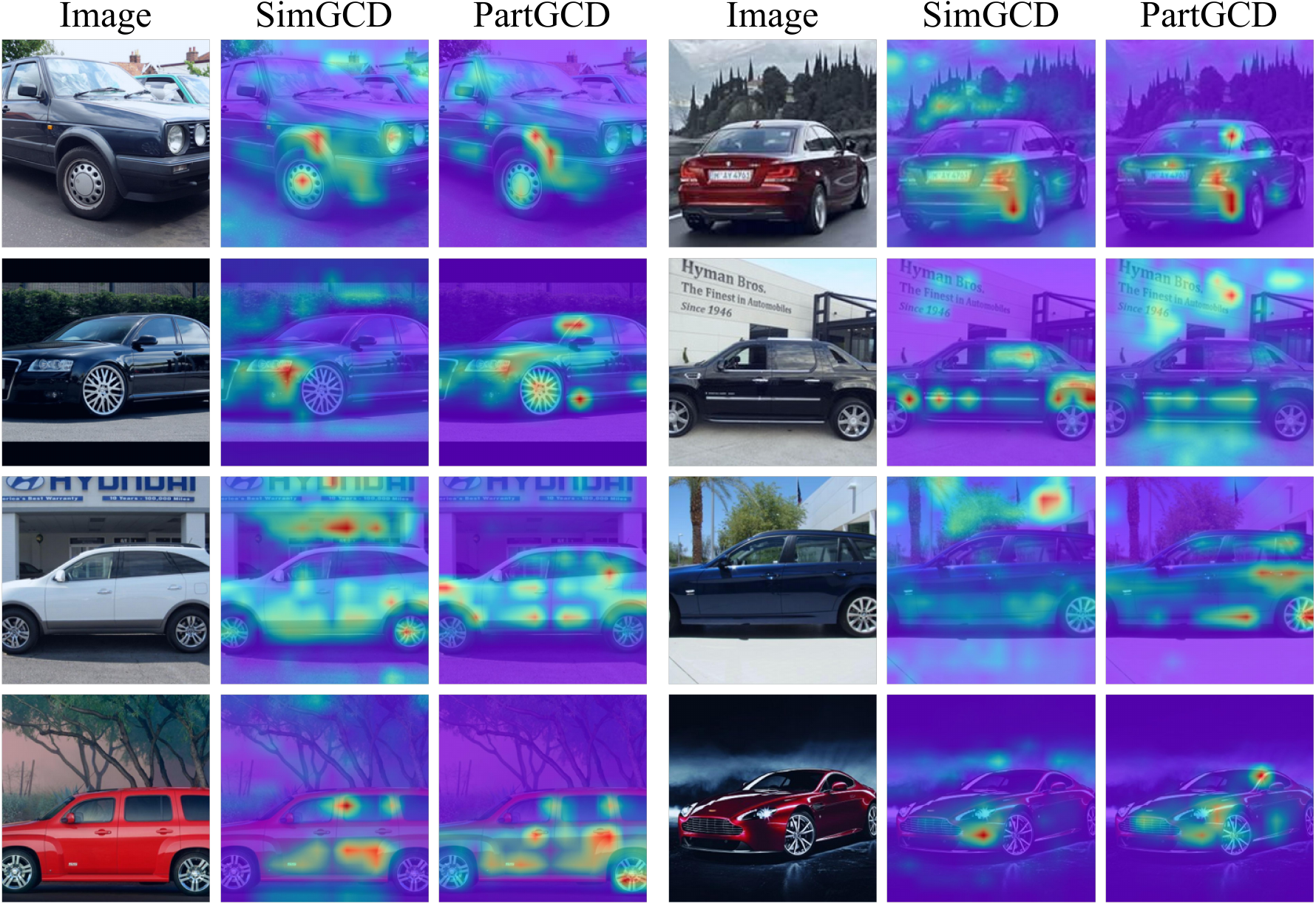}
    \vspace{-5mm}
    \caption{ Attention map visualization.
    }
    \label{fig:car_attn}
\end{figure}
\paragraph{ Attention Map Visualization.}
We show the class-to-patch token attention average across all heads for some instances from the SCars dataset in~\cref{fig:car_attn}.
Our method integrates part knowledge with global knowledge through shared parameters using a parametric training strategy, enabling communication between local and global information. Therefore, our attention maps tend to be more locally aware and more accurate, which demonstrates our method's better understanding of categories and establishes a better visual-semantic relationship.

\begin{figure}[!h]
    \centering
    \includegraphics[width=\columnwidth]{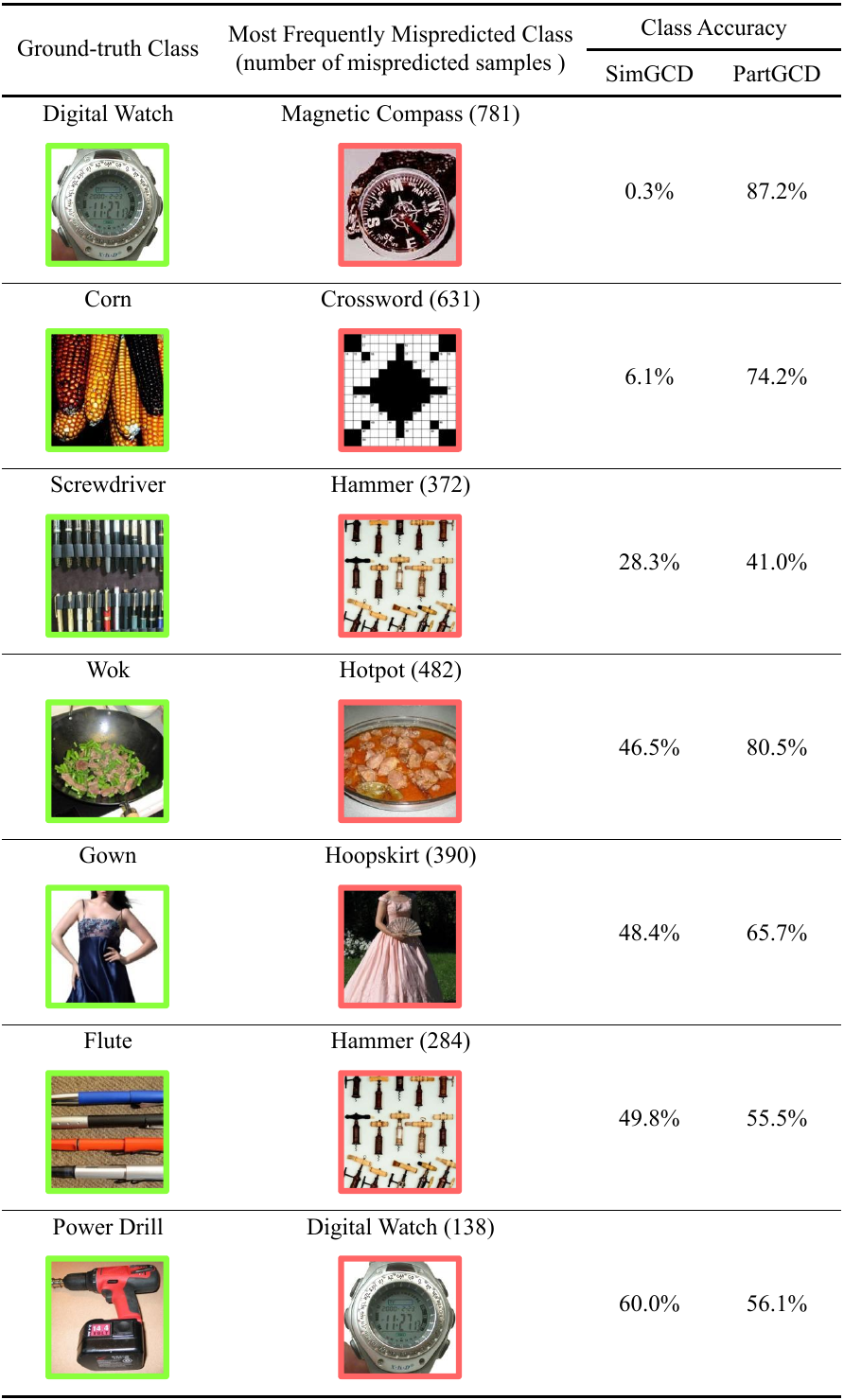}
    \vspace{-5mm}
    \caption{Results on visually similar classes in ImageNet-100. We present the top six categories with the lowest classification accuracy in SimGCD, observing that these categories are often misclassified as other visually similar classes. Our improved results highlight the effectiveness of generic datasets.
    We also provide a failure case in the final row.
    }
    \label{fig:in100sim}
\end{figure}

\paragraph{Results on Visually Similar Classes in ImageNet-100.}
As mentioned in the main paper, our method demonstrates its effectiveness on generic datasets by distinguishing visually similar classes. \cref{fig:in100sim} presents the top six categories with the lowest classification accuracy in SimGCD on the ImageNet-100 dataset. We find that these classes are easily mispredicted to other visually similar classes, challenging the model’s ability to differentiate them. Our improved results highlight PartGCD’s effectiveness on generic datasets, even though it was designed for fine-grained GCD tasks.

\begin{figure*}[!t]
    \centering
    \includegraphics[width=\textwidth]{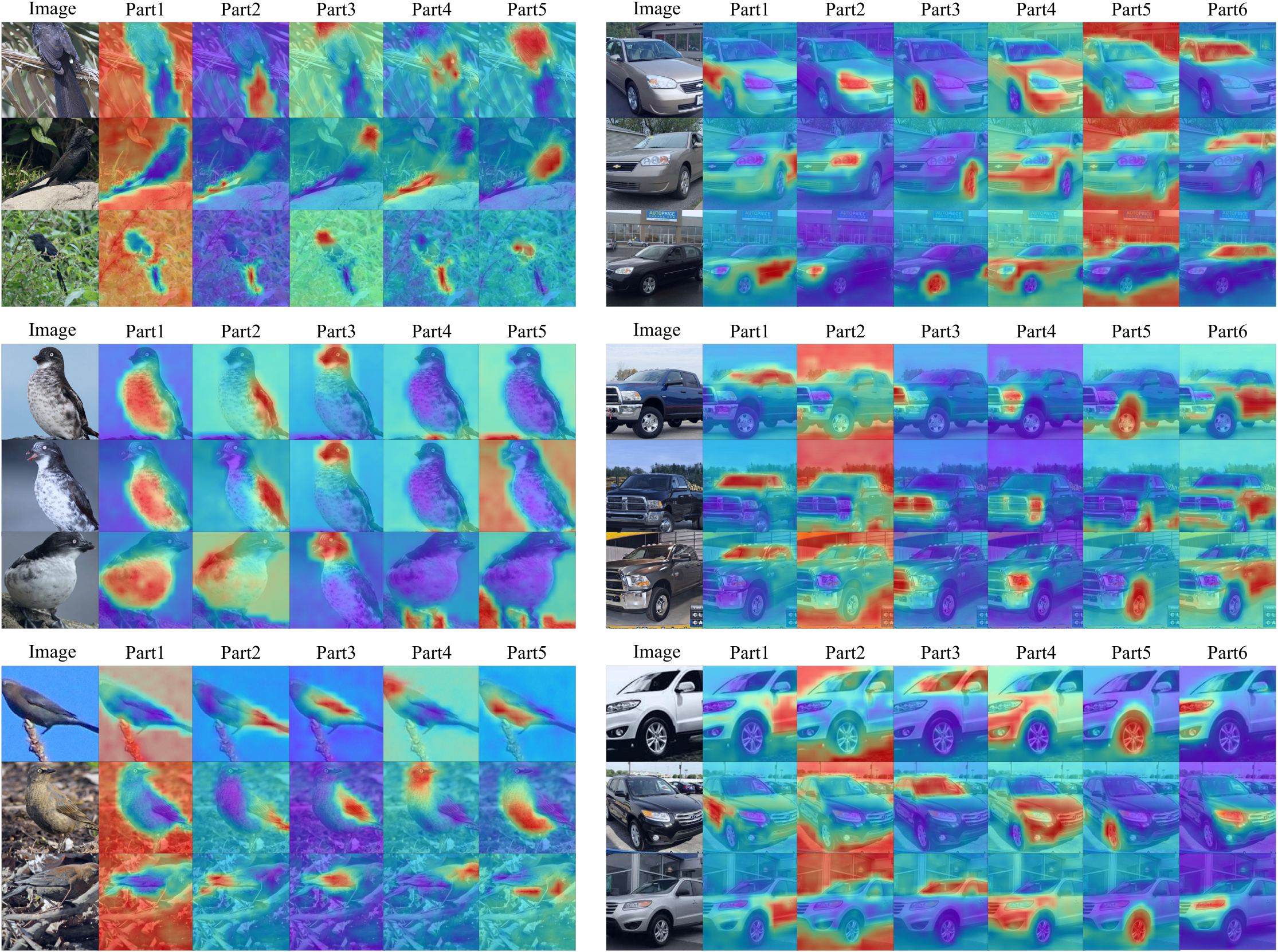}
    \caption{Additional part attention maps.
    }
    \label{fig:more_pa}
\end{figure*}

\paragraph{ Additional Part Attention Maps Visualization for SCars and CUB Datasets.}
In~\cref{fig:more_pa}, we present
additional part attention maps for the
SCars and CUB datasets, showing our effectiveness in generating part attention maps.

\newpage

\end{document}